\documentclass[lettersize,journal]{IEEEtran}
\usepackage{amsmath,amsfonts}
\usepackage{algorithmic}
\usepackage{algorithm}
\usepackage{array}
\usepackage[caption=false,font=normalsize,labelfont=sf,textfont=sf]{subfig}
\usepackage{textcomp}
\usepackage{stfloats}
\usepackage{url}
\usepackage{verbatim}
\usepackage{graphicx}
\usepackage{multirow}
\usepackage{cite}
\usepackage{amsmath}
\usepackage{booktabs}
\usepackage{xcolor}
\usepackage{ulem}

\newcommand{\tabincell}[2]{\begin{tabular}{@{}#1@{}}#2\end{tabular}}

\begin{document}

\title{A Comprehensive Survey on Underwater Image Enhancement Based on Deep Learning}

\author{Xiaofeng Cong, Yu Zhao, Jie Gui, Junming Hou, Dacheng Tao
        \thanks{
            (X. Cong and Y. Zhao contributed equally to this work.) (Corresponding author: J. Gui.)

                X. Cong and Y. Zhao are with the School of Cyber Science and Engineering, Southeast University,  Nanjing 210000, China (e-mail: cxf\_svip@163.com, zyzzustc@gmail.com).

                J. Gui is with the School of Cyber Science and Engineering, Southeast University and with Purple Mountain Laboratories, Nanjing,  and with Engineering Research Center of Blockchain Application, Supervision And Management (Southeast University), Ministry of Education, 210000, China (e-mail: guijie@seu.edu.cn).

                J. Hou is with the State Key Laboratory of Millimeter Waves, School of Information Science and Engineering, Southeast University, Nanjing 210096, China (e-mails: junming\_hou@seu.edu.cn).

                D. Tao is with the College of Computing \& Data Science at Nanyang Technological University, \#32 Block N4 \#02a-014, 50 Nanyang Avenue, Singapore 639798 (email: dacheng.tao@gmail.com).
                
}
}

\markboth{Journal of \LaTeX\ Class Files,~Vol.~14, No.~8, August~2021}%
{Shell \MakeLowercase{\textit{et al.}}: A Sample Article Using IEEEtran.cls for IEEE Journals}


\makeatletter
\def\ps@IEEEtitlepagestyle{
  \def\@oddfoot{\mycopyrightnotice}
  \def\@evenfoot{}
}
\def\mycopyrightnotice{
  {\footnotesize
  \begin{minipage}{\textwidth}
  \centering
  Copyright~\copyright~2025 IEEE. Personal use of this material is permitted.  Permission from IEEE must be obtained for all other uses, in any current or future media, including reprinting/republishing this material for advertising or promotional purposes, creating new collective works, for resale or redistribution to servers or lists, or reuse of any copyrighted component of this work in other works.
  \end{minipage}
  }
}

\maketitle

\begin{abstract}
    Underwater image enhancement (UIE) presents a significant challenge within computer vision research. Despite the development of numerous UIE algorithms, a thorough and systematic review is still absent. To foster future advancements, we provide a detailed overview of the UIE task from several perspectives. Firstly, we introduce the physical models, data construction processes, evaluation metrics, and loss functions. Secondly, we categorize and discuss recent algorithms based on their contributions, considering six aspects: network architecture, learning strategy, learning stage, auxiliary tasks, domain perspective, and disentanglement fusion. Thirdly, due to the varying experimental setups in the existing literature, a comprehensive and unbiased comparison is currently unavailable. To address this, we perform both quantitative and qualitative evaluations of state-of-the-art algorithms across multiple benchmark datasets. Lastly, we identify key areas for future research in UIE. A collection of resources for UIE can be found at {https://github.com/YuZhao1999/UIE}.
\end{abstract}


\section{Introduction}
\begin{figure*}
    \small
    \centering
    \includegraphics[width=17.7cm,height=2.8cm]{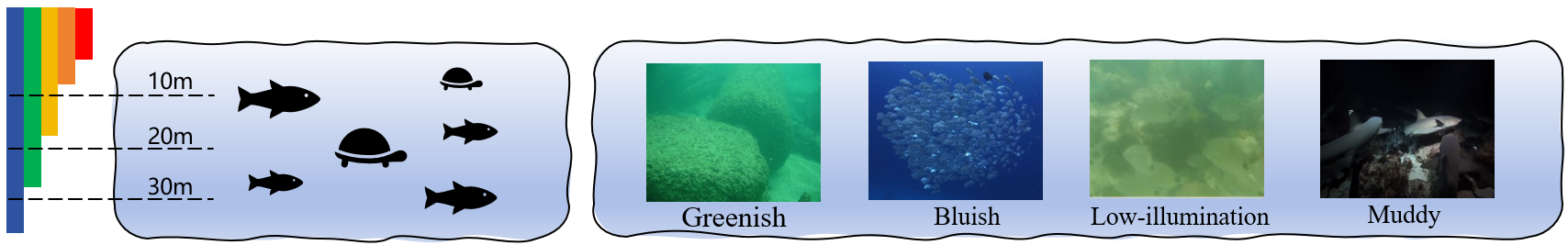}
    \leftline{\hspace{0.5cm} (a) Underwater distortion \cite{zhang2019survey,zhou2023underwater,wang2019experimental} \hspace{4.5cm} (b) Various distortions}
    \caption{Underwater degradation.}
    \label{fig:demo_underwater_degradation}
  \end{figure*}

\IEEEPARstart{U}{nderwater} imaging is an important task in the field of computer vision \cite{ancuti2017color,wang2024metalantis}. High-quality underwater images are necessary for underwater resource exploration, film shooting, personal entertainment, etc. However, due to the absorption and scattering effects of light in underwater scenes, the quality of underwater images may be degraded to varying degrees \cite{chiang2011underwater}. As shown in Fig. \ref{fig:demo_underwater_degradation}-(a), as the depth of water increases, the red, orange, yellow and green light components disappear in sequence. Meanwhile, as shown in Fig. \ref{fig:demo_underwater_degradation}-(b), diverse distortions may occur in underwater environments. Common types of distortion are as follows.
\begin{itemize}
    \item \textit{Color distortion and contrast reduction}: Due to the different attenuation degrees of light with different wavelengths, the color of  images tends to be blue-green \cite{gao2019underwater}.
    \item \textit{Hazy, noisy and blurry effect}: Light may be significantly attenuated in water, which is usually caused by suspended particles or muddy water. The image obtained in such environment may appear hazy, noisy or blurry \cite{wang2021leveraging}.
    \item \textit{Low illumination}: When the depth of water exceeds a certain value, the environment lighting approaches a low-light state, which requires an auxiliary light source. \cite{hou2023non}.
\end{itemize}

Aiming at improving the quality of degraded underwater images collected in complex underwater environments, various \underline{u}nderwater \underline{i}mages \underline{e}nhancement (UIE) methods have been proposed. Existing algorithms can be divided into \underline{n}on-\underline{d}eep \underline{l}earning-based UIE (NDL-UIE) and \underline{d}eep \underline{l}earning-based UIE (DL-UIE). Various prior assumptions, physical models and non-data-driven classical image processing methods are widely utilized by NDL-UIE \cite{song2020enhancement,drews2016underwater,song2018rapid,zhang2022underwater,zhuang2022underwater}. However, due to the complexity of the underwater environment, the strategy adopted by NDL-UIE may be inaccurate in certain scenarios. Problems include (i) the lack of perfect physical modeling for underwater scenarios, (ii) inherent errors in the estimation of physical parameters, (iii) the possibility that specific prior assumptions may not be applicable in every scene, and (iv) the limitation that classical image processing methods do not accommodate scene-specific representations \cite{yang2019depth,anwar2020diving,raveendran2021underwater,jian2021underwater}.

Considering the impressive performance of data-driven algorithms in the field of computer vision and image processing, learning-based solutions have received more attention from researchers. Aiming at further improving the performance of the UIE task, various DL-UIE algorithms have been proposed and verified \cite{yan2022attention,islam2020fast,chen2021underwater,fu2022underwater,naik2021shallow,wang2022semantic,fabbri2018enhancing,liu2019underwater,li2019underwater,qi2022sguie,huang2023contrastive}. Focusing on the problems faced by the UIE task, various efforts are expended by researchers. To promote future research, we summarize the UIE task from multiple perspectives, including
\begin{itemize}
    \item Proposed solutions. 
    \item Results achieved by existing solutions.
    \item Issues worthy of further exploration.
\end{itemize}

\textit{Solutions have been proposed.} On the one hand, we summarize the physical models, data generation methods, and evaluation metrics which are constructed to serve as the basis for research. On the other hand, numerous network structures and training strategies are classified and discussed. In order to know which methods have been widely explored, we need to understand the similarities and differences between them. Within our perspective, existing research can be subjectively viewed as being carried out from six different aspects, namely network architecture, learning strategy, learning stage, assistance task, domain perspective and disentanglement $\&$ fusion. To facilitate future research that can easily build on existing work, we provide a comprehensive review of existing algorithms. According to the main contributions of different algorithms, we divide the existing work into 6 first-level categories. Then, for each first-level category, we give the corresponding second-level category. The taxonomy of algorithms is shown in Table \ref{tab:taxonomy_UIE_methods}. 

\textit{The results achieved by existing solutions.} In existing papers, \underline{s}tate-\underline{o}f-\underline{t}he-\underline{a}rt (SOTA) algorithms have been verified on benchmark datasets. Meanwhile, their performance has been compared with other algorithms. However, the experimental settings adopted in different papers may be inconsistent. A comprehensive and fair comparative experiment for the UIE task has not yet been conducted. Here, we provide a fair comparison by unifying various experimental settings with the aim of obtaining answers on the following two questions. 
\begin{itemize}
    \item Under fair settings, how do SOTA algorithms perform?
    \item Which algorithms achieves the best performance?
\end{itemize}

\textit{Issues worthy of further exploration.} Based on our systematic review of existing algorithms and comprehensive experimental evaluation, we further discuss issues worthy of future research. Specifically, high-quality data synthesis, cooperation with text-image multi-modal models, non-uniform illumination, reliable evaluation metrics, and combination with other image restoration tasks are topics that still need to be studied in depth. In a word, the UIE is in an emerging stage, rather than a task that has been almost perfectly solved.

\subsection{The difference between this survey and others}
The \cite{zhang2019survey,yang2019depth,raveendran2021underwater,jian2021underwater,wang2019experimental} mainly introduce the traditional UIE algorithms, while the latest DL-UIE algorithms are rarely mentioned. Anwar et al. \cite{anwar2020diving} classify DL methods from the perspective of network architecture. However, the latest progress in the research of the UIE task such as Fourier operation \cite{wei2022uhd}, contrastive learning \cite{liu2022twin}, and rank learning \cite{guo2023underwater} are not mentioned. The perspective of \cite{zhou2023underwater} is the difference between hardware-based and software-based algorithms, so a comprehensive discussion of DL-based algorithms is not the goal of \cite{zhou2023underwater}. In this survey, we provide a comprehensive discussion of recent DL-based advancements.

\subsection{A guide for reading this survey}
The terminology in the existing literature may not be consistent. To facilitate the reading of this paper, especially the figures, we summarize the important terminologies and their meanings as follows.
\begin{itemize}
    \item \textbf{Degradation}: Quality degradation during underwater imaging due to absorption and scattering phenomena.
    \item \textbf{Distortion $x$}: The degraded quality underwater image taken in different types of water.
    \item \textbf{Reference $y$}: The underwater image with subjectively higher quality. There is no such thing as a perfect, distortion-free underwater image.
    \item \textbf{Prediction $\hat{y}$}: A visual quality enhanced underwater image obtained by an UIE algorithm.
    \item \textbf{Paired images}: A distortion image $x$ with the corresponding reference image $y$.
    \item \textbf{In-air images}: Images taken in a terrestrial scene, which may be indoors or outdoors.
\end{itemize}

The terminology used for the task of improving underwater image quality has certain differences in existing literatures, which includes underwater image enhancement, underwater image restoration and underwater image dehazing. Since certain similarities exist between the task we studied and the image dehazing task, such as similar imaging principles \cite{carlevaris2010initial}, some literature uses ``underwater image dehazing'' as a definition. This definition has been used less frequently in recent literature \cite{anwar2020diving}. Furthermore, ``enhancement'' and ``restoration'' are often used in different literature. There is no significant difference between these two terms in the task of improving underwater image quality. Therefore, in this survey, we use ``enhancement'' uniformly to define the task we discuss.

For ease of reading, abbreviations are used to refer to each UIE algorithm mentioned in the paper. Many papers provide abbreviations for their algorithms, such as URanker \cite{guo2023underwater}. For papers that did not provide an algorithm abbreviation, we constructed the abbreviation using the first letters of the words in the title that illustrate their main contributions.

The commonly used physical models, data generation methods, evaluation metrics and loss functions are summarized in Section \ref{sec:related_work}. The taxonomy and analysis of existing algorithms are placed in Section \ref{sec:taxonomy}. Section \ref{sec:experiments} provides comprehensive experiments and summarizes the conclusions. Challenging and valuable issues that have not yet been solved are discussed in Section \ref{sec:future_work}. A summary of this paper is in Section \ref{sec:conclusion}.

\begin{table*}
    \renewcommand{\arraystretch}{1.3}
    \centering
    \caption{A taxonomy of UIE algorithms.}
    \label{tab:taxonomy_UIE_methods}
    \begin{tabular}{ccc}
    \hline
    Category & Key Idea & Methods \\
     
    \hline
    \multirow{6}{*}{\tabincell{c}{Network Architecture}} & Convolution Operation & \tabincell{c}{UWCNN \cite{li2020underwater}, UWNet \cite{naik2021shallow}, UResNet \cite{liu2019underwater}, DUIR \cite{dudhane2020deep}, \\ UIR-Net \cite{mei2022uir}, FloodNet \cite{gangisetty2022floodnet}, LAFFNet \cite{yang2021laffnet}, UICoE-Net \cite{qi2021underwater}, PUIE-Net \cite{fu2022uncertainty}} \\

    \cline{2-3}
    & Attention Mechanism      & \tabincell{c}{WaveNet \cite{sharma2023wavelength}, ADMNNet \cite{yan2022attention}, LightEnhanceNet \cite{zhang2024liteenhancenet}, \\ CNMS \cite{zhang2024robust}, HAAM-GAN \cite{zhang2023hierarchical}, MFEF \cite{zhou2023multi}} \\

    \cline{2-3}
    & Transformer Module & \tabincell{c}{PTT \cite{boudiaf2022underwater}, U-Trans \cite{peng2023u}, UWAGA \cite{huang2022underwater}, WaterFormer \cite{wen2024waterformer}, Spectroformer \cite{khan2024spectroformer}} \\

    \cline{2-3}
    & Fourier Transformation        & \tabincell{c}{UHD \cite{wei2022uhd}, WFI2-Net \cite{zhao2023wavelet}, TANet \cite{zhang2024tanet}, UIE-INN \cite{chu2023underwater}, SFGNet \cite{zhao2023toward}} \\

    \cline{2-3}
    & Wavelet Decomposition         & \tabincell{c}{UIE-WD \cite{ma2022wavelet}, EUIE \cite{jamadandi2019exemplar},  PRWNet \cite{huo2021efficient}, MWEN \cite{wang2024multi}} \\

    \cline{2-3}
    & Neural Architecture Search & \tabincell{c}{AutoEnhancer \cite{tang2022autoenhancer}} \\
    
    \hline
    \multirow{5}{*}{\tabincell{c}{Learning Strategy}} & Adversarial Learning     & \tabincell{c}{DGD-cGAN \cite{gonzalez2024dgd}, FGAN \cite{li2019fusion}, UIE-cGAN \cite{yang2020underwater}, TOPAL \cite{jiang2022target}, \\ FUnIEGAN \cite{islam2020fast}, EUIGAN \cite{fabbri2018enhancing}, CE-CGAN \cite{agarwal2023contrast}, RUIG \cite{desai2021ruig}} \\

    \cline{2-3}
    & Rank Learning          & \tabincell{c}{URanker \cite{guo2023underwater}, PDD-Net \cite{jiang2023perception}, CLUIE-Net \cite{li2022beyond}} \\

    \cline{2-3}
    & Contrastive Learning     & \tabincell{c}{TACL \cite{liu2022twin}, Semi-UIR \cite{huang2023contrastive}, CWR \cite{han2021single}, DRDCL \cite{yin2024unsupervised}, TFUIE \cite{yu2023task}, \\ RUIESR \cite{li2023ruiesr}, CL-UIE \cite{tian2023deformable}, HCLR-Net \cite{zhou2024hclr}, UIE-CWR \cite{han2022underwater}} \\

    \cline{2-3} 
    & Reinforcement Learning   & \tabincell{c}{HPUIE-RL \cite{song2024hierarchical}} \\

    \hline
    \multirow{3}{*}{\tabincell{c}{Learning Stage}} & Single Stage     & UWCNN \cite{li2020underwater}, UWNet \cite{naik2021shallow}, UResNet \cite{liu2019underwater} \\
    \cline{2-3}
    & Coarse-to-fine  & \tabincell{c}{MBANet \cite{xue2023investigating}, GSL \cite{lin2022global}, CUIE \cite{khandouzi2024coarse}, DMML \cite{esmaeilzehi2024dmml}} \\

    \cline{2-3}
    & Diffusion Learning       & \tabincell{c}{UIE-DM \cite{tang2023underwater}, SU-DDPM \cite{lu2023speed}, CPDM \cite{shi2024cpdm}
    } \\

    \hline
    \multirow{2}{*}{\tabincell{c}{Assistance Task}} & Semantic Assistance     & SGUIE \cite{qi2022sguie}, SATS \cite{wang2022semantic}, DAL-UIE \cite{kapoor2023domain}, WaterFlow \cite{zhang2023waterflow}, HDGAN \cite{chen2020perceptual} \\

    \cline{2-3}
    & Depth Assistance         & \tabincell{c}{Joint-ID \cite{yang2023joint}, DAUT \cite{badran2023daut}, DepthCue \cite{desai2023depthcue}, HybrUR \cite{yan2023hybrur}} \\

    \hline
    \multirow{3}{*}{\tabincell{c}{Domain Perspective}}    & Knowledge Transfer          & \tabincell{c}{TUDA \cite{wang2023domain}, TSDA \cite{jiang2022two}, DAAL \cite{zhou2020domain}, IUIE \cite{bing2023domain}, \\ WSDS-GAN \cite{liu2023wsds}, TRUDGCR \cite{jain2022towards}, UW-CycleGAN \cite{yan2023uw}} \\

    \cline{2-3} 
    & Domain Translation       & \tabincell{c}{URD-UIE \cite{zhu2023unsupervised}, TACL \cite{liu2022twin}} \\

    \cline{2-3}
    & Diversified Output       & \tabincell{c}{UIESS \cite{chen2022domain}, UMRD \cite{zhu2023unsupervised_2}, PWAE \cite{kim2021pixel}, CECF \cite{cong2024underwater}} \\

    \hline
    \multirow{5}{*}{\tabincell{c}{Disentanglement $\&$ Fusion}}     & Physical Embedding       & \tabincell{c}{IPMGAN \cite{liu2021ipmgan}, ACPAB \cite{lin2020attenuation}, USUIR \cite{fu2022unsupervised}, AquaGAN \cite{desai2022aquagan}, PhysicalNN \cite{chen2021underwater}, GUPDM \cite{mu2023generalized}} \\ 

    \cline{2-3}
    & Retinex Model            & \tabincell{c}{CCMSR-Net \cite{qi2023deep}, ReX-Net \cite{zhang2023rex}, ASSU-Net \cite{khan2023underwater}} \\

    \cline{2-3} 
    & Color Space Fusion       & \tabincell{c}{UIEC$^2$-Net \cite{wang2021uiec}, UGIF-Net \cite{zhou2023ugif}, TCTL-Net \cite{li2023tctl}, MTNet \cite{moran2023mtnet}, UDNet \cite{saleh2022adaptive}, \\
    P2CNet \cite{rao2023deep}, JLCL-Net \cite{xue2021joint}, TBDNN \cite{hu2021two}, MSTAF \cite{ji2024real}, UColor \cite{li2021underwater}} \\

    \cline{2-3} 
    & Water Type Focus         & \tabincell{c}{SCNet \cite{fu2022underwater}, DAL \cite{uplavikar2019all}, IACC \cite{zhou2024iacc}} \\       

    \cline{2-3}
    & Multi-input Fusion       & \tabincell{c}{WaterNet \cite{li2019underwater}, MFEF \cite{zhou2023multi}, F2UIE \cite{verma2023f2uie}} \\

    \hline
    \end{tabular}
    \end{table*}

\section{Related Work}
\label{sec:related_work}
In this section, the six aspects involved in the UIE task are introduced. The definition of UIE research and the scope of this paper are given in \ref{subsec:problem_definition}. Similar problems faced by the UIE task and other image restoration tasks are discussed in \ref{subsec:connection_uie_with_other_restoration_tasks}. Physical models that are considered reliable are presented in Section \ref{subsec:physical_models}. Section \ref{subsec:datasets} summarizes common data generation and collection methods. The widely used evaluation metrics and loss functions are given in Section \ref{subsec:evaluation_metrics} and Section \ref{subsec:loss_functions}, respectively.

\subsection{Problem Definition and Research Scope}
\label{subsec:problem_definition}
Following the way shown in \cite{li2021low}, we give a commonly used definition of the DL-UIE. For a given distorted underwater image $x$, the enhancement process can be regarded as
\begin{equation}
    \hat{y} = \Phi(x; \theta),
\end{equation}
where $\Phi$ denotes the neural network with the learnable parameters $\theta$. Both $x$ and $\hat{y}$ belong to $\mathcal{R}^{H \times W \times 3}$. For UIE models, a common optimization purpose is to minimize the error
\begin{equation}
    \hat{\theta} = \arg \min \mathcal{L}(\hat{y}, y),
\end{equation}
where $\mathcal{L}(\hat{y}, y)$ denotes the loss function for obtaining the optimal parameters $\hat{\theta}$. $\mathcal{L}(\hat{y}, y)$ may be supervised or unsupervised loss functions with any form.

Based on the above definition, it is worth clearly pointing out that our survey is aimed at the investigation of DL-UIE algorithms with single-frame underwater images as input. Therefore, algorithms based on multi-frame images \cite{xie2024uveb} or that do not utilize deep learning at all \cite{li2016single,peng2017underwater,carlevaris2010initial} are outside the scope of this paper.

\subsection{The connections between the UIE task and other image restoration tasks}
\label{subsec:connection_uie_with_other_restoration_tasks}
The UIE task is regarded as a sub-research field of low-level image restoration tasks. Similar problems arise in the UIE task and other image restoration tasks. The common faced issues include, (i) hazy effect caused by absorption and scattering, like image dehazing \cite{han2018review}, (ii) blurry details caused by camera shake, light scattering, or fast target motion, like deblurring \cite{zhang2022deep}, (iii) noise caused by suspended particles \cite{jian2021underwater,jiang2020novel}, like image denoising \cite{chen2023masked}, (iv) low-illumination due to poor light, like low-light image enhancement \cite{xu2023low}, (v) non-uniform illumination caused by artificial light sources, like nighttime dehazing \cite{liu2022multi}. In general, the problems faced by underwater imaging may be regarded as a complex combination of multiple image restoration tasks.

\subsection{Physical Models}
\label{subsec:physical_models}
To the best of our knowledge, there is currently no model that perfectly describes the underwater imaging process \cite{anwar2020diving}, which is totally suitable for algorithmic solution. Here, two widely used and proven effective imaging models are introduced, namely the atmospheric scattering model \cite{drews2013transmission} and revised underwater image formation model \cite{akkaynak2018revised}. The atmospheric scattering model \cite{anwar2020diving} is widely adopted in the research of image dehazing and underwater image enhancement \cite{ye2022underwater}, which can be expressed as
\begin{equation}
    x = y \cdot e^{-\beta d} + B^{\infty} \cdot (1 - e^{-\beta d}),
\end{equation}
where $d$, $B^{\infty}$ and $\beta$ denote the scene depth, background light and attenuation coefficient, respectively. The revised underwater image formation model \cite{akkaynak2018revised}, which is defined as
\begin{equation}
    x = y e^{-\beta^{D}(v_{D}) d} + B^{\infty} \left(1 - e^{\beta^{B}(v_{B}) d}\right),
\end{equation}
where $\beta^{B}$ and $\beta^{D}$ represent the backscatter and direct transmission attenuation coefficients \cite{anwar2020diving}. The $v_{B}$ and $v_{D}$ are the dependent coefficients for $\beta^{B}$ and $\beta^{D}$. The atmospheric scattering model  and revised underwater image formation model can be used for synthesis of data and design of algorithms. Since an in-depth discussion of the imaging mechanism may be beyond the scope of this paper, more details on the physical model can be found in \cite{akkaynak2018revised,akkaynak2017space,akkaynak2019sea}.


\subsection{Datasets for Model Training and Performance Evaluation}
\label{subsec:datasets}

Pairs of real-world distorted underwater images and reference images are difficult to obtain \cite{hambarde2021uw,li2019fusion}. The data used for the training and evaluation processes need to be collected by various reliable ways. We group existing ways of building datasets into the following categories.

\begin{itemize}
\item \textbf{\textit{Voted} by SOTA UIE algorithms.} Twelve algorithms are used by UIEB \cite{li2019underwater} to generate diverse reference images. The best enhancement effect for each scene is then selected by a vote of 50 volunteers as the final reference image. Meanwhile, 18 UIE algorithms and 20 volunteers are involved in the voting process of LSUI's \cite{peng2023u} reference images. SAUD \cite{jiang2022underwater} proposes a subjectively annotated benchmark that contains both real-world raw underwater images and available ranking scores under 10 different UIE algorithms.

\item \textbf{\textit{Generated} by domain transformation algorithms.} EUVP \cite{islam2020fast} and UFO-120 \cite{islam2020simultaneous} treat the unpaired distorted image and the reference image as two domains. Pairs of images in EUVP and UFO-120 are generated via an unsupervised domain transformation algorithm \cite{zhu2017unpaired} with cycle consistency constraint.

\item \textbf{\textit{Rendered} by light field.} UWNR \cite{ye2022underwater} designs a light field retention mechanism to transfer the style from natural existing underwater images to objective generated images. UWNR adopts a data-driven strategy, which means that it may avoid the accuracy limitations caused by the complexity of the physical model.

\item \textbf{\textit{Collected} by the undersea image capturing system.} By setting up a multi-view imaging system under seawater, RUIE \cite{liu2020real} constructs an underwater benchmark under natural light. The scenes in RUIE show a natural marine ecosystem containing various sea life, such as fish, sea urchins, sea cucumbers and scallops.

\item \textbf{\textit{Synthesized} by physical model.} Physical models are used by SUID \cite{hou2020benchmarking}, RUIG \cite{desai2021ruig} and WaterGAN \cite{li2017watergan} to estimate the parameters that describe the imaging process.
\end{itemize}

In addition to the methods introduced above, neural radiance field is used by a recent work to synthesize both degraded and clear multi-view images \cite{zhou2023waterhe}. Currently, there is no reliable evidence on which way of constructing a dataset is optimal. Despite the impressive advances that have been achieved, existing methods of constructing datasets still have limitations in various aspects. The quality of the label images produced by the voting method is inherently limited, that is, the best performance of the model being evaluated may not exceed the various algorithms used in the voting process. The diversity of attenuation patterns of images obtained by attenuation synthesis processes inspired by generative models may be limited. Since the generation process with cycle consistency is limited by the one-to-one training mode, which results in cross-domain information not being fully taken into account \cite{almahairi2018augmented}. The complexity of underwater imaging prevents physical models from accurately controlling diverse degradation parameters \cite{ye2022underwater}. Effective data acquisition strategies remain a pressing challenge.

\subsection{Evaluation Metrics}
\label{subsec:evaluation_metrics}

The way to evaluate the performance of UIE models is a challenging topic under exploration. Currently, full-reference and no-reference evaluation metrics are the most widely used metrics. In addition, human subjective evaluation, downstream task evaluation and model efficiency are also adopted by various literature. Details are as follows.

\begin{itemize}
\item \textbf{Full-reference metrics.} The Peak Signal-to-Noise Ratio (PSNR) \cite{gui2023comprehensive} and Structural Similarity (SSIM) \cite{wang2004image} are full-reference evaluation metrics that are widely used for datasets with paired images. They can accurately describe the distance between prediction and reference. 

\item \textbf{No-reference metrics.} For the evaluation of real-world enhancement performance without reference images, no-reference metrics are needed. The Underwater Image Quality Measures (UIQM) \cite{panetta2015human} and Underwater Color Image Quality Evaluation (UCIQE) \cite{yang2015underwater} are widely adopted as no-reference metrics in literature. Efforts \cite{zheng2022uif}, \cite{jiang2022underwater} are still being made to develop more reliable metrics for no-reference assessment.

\item \textbf{Human subjective evaluation.} Human vision-friendly enhanced underwater images are one of the purposes of UIE models. On the one hand, UIE-related papers \cite{yan2022attention,islam2020fast,chen2021underwater,fu2022underwater} usually illustrate the superiority of their proposed methods in color correction, detail restoration, and edge sharpening from the perspective of quantitative analysis. Such an evaluation is usually performed by the authors of the paper themselves. On the other hand, multiple people with or without image processing experience are involved in the evaluation of \cite{li2019underwater,guo2023underwater}. They can give subjective ratings to different algorithms. The best enhancement result may be selected \cite{li2019underwater}, or the enhancement results obtained by different algorithms may be ranked \cite{guo2023underwater}.

\item \textbf{Evaluation by downstream tasks.} The UIE task can be used as an upstream task for other image understanding tasks. Due to the inherent difficulties in UIE models evaluation, that is, the inability to obtain perfect ground-truth, the use of downstream tasks to evaluate UIE models has been widely used in the existing literature. Representative downstream tasks include
\begin{itemize}
    \item object detection \cite{zhao2019object} used in \cite{wu2024self,zhou2023ugif},
    \item feature matching \cite{lowe2004distinctive} used in \cite{yang2023joint,zhou2023multi}, 
    \item  saliency detection \cite{islam2022svam} used in \cite{yang2023joint,zhou2024hclr}, 
    \item semantic segmentation \cite{islam2020semantic} used in \cite{sharma2023wavelength,yu2023task}.
\end{itemize}

\item \textbf{Model efficiency}. An UIE model may be used in underwater devices that do not have significant computing power or storage space. Therefore, the computing and storage resources required by the model are important factors for evaluating an UIE model. The commonly adopted metrics include Flops (G), Params (M) and Inference Time (seconds per image) \cite{zhang2024liteenhancenet,qi2023deep,yin2024unsupervised}.

\end{itemize}
\subsection{Loss Functions}
\label{subsec:loss_functions}

Although different UIE models may adopt different training strategies. Effective loss functions are commonly used by UIE models, as follows.

\begin{itemize}
    \item $\mathcal{L}_1$ loss \cite{gonzalez2024dgd,zhou2023multi}, Smooth $\mathcal{L}_1$ loss \cite{zhang2023rex,zhou2023ugif} and $\mathcal{L}_{2}$ loss \cite{desai2023depthcue,sharma2023wavelength} provide the pixel-level constraint, which is used as fidelity loss.
    \item Perceptual loss \cite{islam2020simultaneous,zhang2023rex,zhou2023multi,zhou2024hclr}, also known as content loss, is used to measure the distance between two images in feature space.
    \item SSIM is the evaluation metric for UIE tasks, which is differentiable. Therefore, SSIM loss \cite{zhou2024iacc,desai2023depthcue,sharma2023wavelength} is widely used for structural constraints.
    \item The attenuation of underwater images may be accompanied by the loss of edge information. Therefore, the edge loss \cite{islam2020simultaneous,hambarde2021uw,moran2023mtnet} calculated by the gradient operator is exploited.
    \item Adversarial losses \cite{desai2021ruig,dudhane2020deep,liu2023wsds} can provide domain discriminative capabilities, which may improve the visual quality of the restored image.
\end{itemize}

\section{UIE Methods}
\label{sec:taxonomy}

According to main contributions of each paper, we divide UIE algorithms into six categories, namely (i) Section \ref{subsec:network_architecture} Network Architecture, (ii) Section \ref{subsec:learning_process} Learning Strategy, (iii) Section \ref{subsec:learning_stage} Learning Stage, (iv) Section \ref{subsec:assistance_task} Assistance Task, (v) Section \ref{subsec:domain_perspective} Domain Perspective and (vi) Section \ref{subsec:disentanglement_and_fusion} Disentanglement $\&$ Fusion, respectively. Details are as follows.
\begin{itemize}
    \item \textbf{Network Architecture}. Network architecture is crucial to the performance of UIE models. The construction of UIE models mainly uses convolution, attention \cite{niu2021review}, Transformer \cite{liu2021swin}, Fourier \cite{duhamel1990fast} and Wavelet \cite{huang2017wavelet} operations.
    \item \textbf{Learning Strategy}. Beyond conventional end-to-end supervised training, due to the complexity of UIE scenarios, diverse learning processes are adopted by the UIE algorithm. The typical learning process includes adversarial learning \cite{jabbar2021survey}, rank learning \cite{zhang2021ranksrgan}, contrastive learning \cite{khosla2020supervised} and reinforcement learning \cite{williams1992simple}.
    \item \textbf{Learning Stage}. The mapping from distortion to prediction may be implemented by the single stage, coarse-to-fine, or stepwise diffusion process \cite{ho2020denoising}.
    \item \textbf{Assistance Task}. Semantic segmentation, object detection, or depth estimation tasks are common auxiliary tasks for the UIE task. The training of joint models for different tasks is beneficial to each other.
    \item \textbf{Domain Perspective}. In-air clear images, underwater distorted images, and underwater clear images can all be regarded as independent domains. From a domain perspective, knowledge transfer, degradation conversion, and enhancement effect adjustment can all be achieved through domain-aware training. 
    \item \textbf{Disentanglement $\&$ Fusion}. Considering the degradation properties of underwater images, disentanglement and fusion are used as two effective ways to improve the interpretability of the model. Physical models and Retinex models are usually used as the theoretical basis for disentanglement. Color space fusion, water type fusion and multi-input fusion are common solutions for fusion.
\end{itemize}

As we introduce the algorithms in each category, schematic diagrams are used to illustrate the differences between the different algorithms. It is worth pointing out that what is expressed in the schematic diagram is not the details of a specific algorithm, but the main idea of a type of algorithms.
\subsection{Network Architecture}
\label{subsec:network_architecture}

\begin{figure*} 
    \small
    \centering
    \includegraphics[width=17.7cm,height=3.6cm]{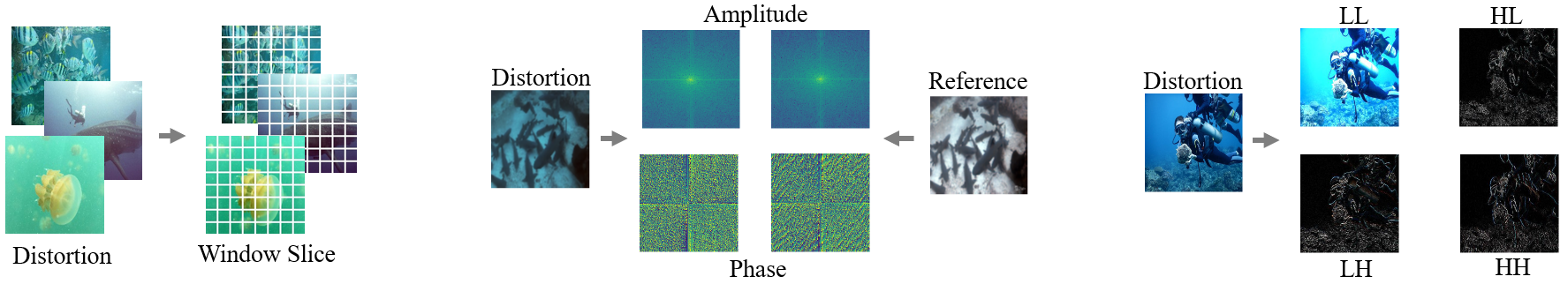}
    \leftline{\hspace{0.3cm} (a) Window-based Transformer \hspace{2cm} (b) Fourier Transformation \hspace{3cm} (c) Wavelet Decomposition \cite{wang2024multi}}
    \caption{Building feature maps with high-quality representation by window-based Transformer, Fourier Transformation and Wavelet Decomposition, respectively.}
    \label{fig:Fourier_Wavelet}
  \end{figure*}

\begin{figure*}
    \small
    \centering
    \includegraphics[width=17.7cm,height=1.5cm]{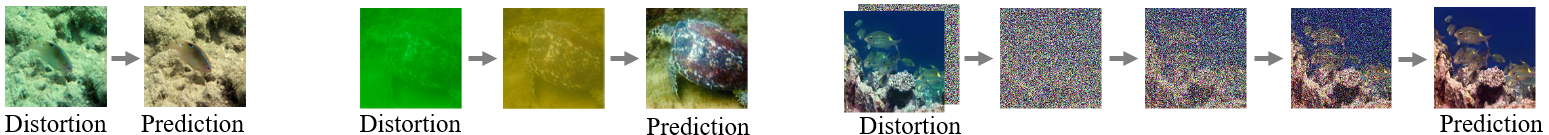}
    \leftline{\hspace{0.5cm} (a) Single Stage \hspace{2cm} (b) Coarse-to-fine \cite{khandouzi2024coarse} \hspace{4cm} (c) Diffusion Process \cite{tang2023underwater}}
    \caption{Generating high-quality enhanced images with varying numbers of stages.}
    \label{fig:singleStage_twoStage_Diffusion}
  \end{figure*}

\subsubsection{Convolution Operation}
Convolution is one of the most commonly used operations in UIE models. Network blocks constructed by convolution operation include naive convolution (UWCNN \cite{li2020underwater}, UWNet \cite{naik2021shallow}), residual connection (UResNet \cite{liu2019underwater}), residual-dense connection (DUIR \cite{dudhane2020deep}, FloodNet \cite{gangisetty2022floodnet}), and multi-scale fusion (LAFFNet \cite{yang2021laffnet}). The computational consumption of UIE models based on convolutional architecture is usually less.

\subsubsection{Attention Mechanism}
In the UIE task, the attention mechanism is mainly used to help the model focus on different spatial locations or channel information. By analyzing the wavelength-driven contextual size relationship of the underwater scene, WaveNet \cite{sharma2023wavelength} assigns operations with different receptive fields to different color channels. Then, attention operations along spatial and channel dimensions are used to enhance convolutional paths with different receptive fields. Based on the selective kernel mechanism, a nonlinear strategy is proposed by ADMNNet \cite{yan2022attention} to change receptive field sizes reasonably by adopting soft attention. Meanwhile, a module with the ability to dynamically aggregate channel information is designed by ADMNNet. Aiming at calibrating the detailed information of the distorted input, MFEF \cite{zhou2023multi} constructs a pixel-weighted channel attention calculation flow.

\subsubsection{Transformer Module}
The Transformer architecture \cite{vaswani2017attention,liu2021swin} has been widely used in natural language processing and computer vision research. UWAGA \cite{huang2022underwater} points out that an automatic selection mechanism for grouping the input channels is beneficial for mining relationships between channels. Based on this view, an adaptive group attention operation embedded in Swin-Transformer \cite{liu2021swin} is designed by UWAGA. As shown in Fig. \ref{fig:Fourier_Wavelet}-(a), under the principle of Swin-Transformer, the image (or feature map) can be split into window-based patches. Image Transformers pretrained on ImageNet are used by PTT \cite{boudiaf2022underwater} to fine-tune on the UIE task, which shows that features with long-distance dependencies are effective for handling the underwater distortion. From both the channel and spatial perspectives, a multi-scale feature fusion transformer block and a global feature modeling transformer block are proposed by U-Trans \cite{peng2023u}. A multi-domain query cascaded Transformer architecture is designed by Spectroformer \cite{khan2024spectroformer}, where localized transmission representation and global illumination information are considered. Although these Transformer-based architectures have proven beneficial for the UIE task, they often incur computational costs that exceed conventional convolution operations.

\subsubsection{Fourier Transformation}
The Fourier transform \cite{chu2023underwater} provides a perspective for analyzing features in the frequency domain. In the UIE task, operations in the Fourier domain and the spatial domain are often used jointly. The Fourier operation for a 2D image signal $x \in \mathcal{R}^{H \times W}$ is
\begin{equation}
    \mathcal{F}(x)(u, v) = \frac{1}{\sqrt{HW}} \sum_{h=0}^{H-1} \sum_{w=0}^{W-1} x(h, w) e^{-j2{\pi}(\frac{h}{H}u + \frac{w}{W}v)},
\end{equation}
where $(h, w)$ and $(u, v)$ denote coordinates in spatial and Fourier domain, respectively. The process of decomposing an underwater image into an amplitude spectrum and a phase spectrum is visualized in Fig. \ref{fig:Fourier_Wavelet}-(b). A dual-path enhancement module that jointly performs feature processing in the spatial and Fourier domains was designed by UHD \cite{wei2022uhd}. In the Fourier domain, channel mixers are used by UHD to extract the frequency domain features of the distorted image. Specifically, the real part and the imaginary part are the objects processed by UHD feature extraction. In the spatial domain, a contracting path and an expansive path are adopted to construct the spatial feature descriptor. Finally, the features in the frequency domain and spatial domain are fused. TANet \cite{zhang2024tanet} designs an atmospheric light removal Fourier module by utilizing the feature information of the real and imaginary parts. WFI2-Net \cite{zhao2023wavelet} and SFGNet \cite{zhao2023toward} also adopt the feature extraction and fusion method in the space-frequency domain. Unlike UHD and TANet, which process the real and imaginary parts, WFI2-Net and SFGNet dynamically filter the amplitude spectrum and phase spectrum. Although Fourier operations have been widely used by the UIE task, to the best of our knowledge, in-depth observations of degradation information in the frequency domain have not been given.

\subsubsection{Wavelet Decomposition}
The description of different frequency sub-bands can be obtained by wavelet decomposition. Four frequency bands can be obtained from the original 2D signal $x$ by Discrete Wavelet Transform (DWT) as 
\begin{equation}
    I_{LL}, I_{LH}, I_{HL}, I_{HH} = DWT(x),
\end{equation}
where $I_{LL}, I_{LH}, I_{HL}, I_{HH}$ denote the low-low, low-high, high-low and high-high sub-bands, respectively. The process of Wavelet decomposition of underwater images is visualized in Fig. \ref{fig:Fourier_Wavelet}-(c). UIE-WD \cite{ma2022wavelet} uses a dual-branch network to process images of different frequency sub-bands separately, in which each branch handles color distortion and detail blur problems respectively. By analyzing the limitations of a single pooling operation, wavelet pooling and unpooling layers based on Haar wavelets are proposed by EUIE \cite{jamadandi2019exemplar}. PRWNet \cite{huo2021efficient} uses wavelet boost learning to obtain low-frequency and high-frequency features. PRWNet proposes that high-frequency sub-bands represent texture and edge information, while low-frequency sub-bands contain color and lighting information. By using normalization and attention, the information of different sub-bands can be refined. A frequency subband-aware multi-level interactive wavelet enhancement module is designed by MWEN \cite{wang2024multi}, aiming at building a feature extraction branch with more expressive capabilities.

\subsubsection{Neural Architecture Search} 
The overall network architecture and the way that modules are connected are important components of the UIE algorithm. Effective network design solutions require designers to have extensive experience and knowledge. Not only that, a large number of tentative experiments need to be performed repeatedly. Therefore, neural architecture search (NAS) is explored by AutoEnhancer \cite{tang2022autoenhancer} for building UIE network structures with impressive performance. An encoder-decoder architecture is chosen by AutoEnhancer as the supernet. Three processes are used as the basis of AutoEnhancer, namely supernet training, subnet searching and subnet retraining. The first step is to obtain the parameter $W^{*}$
\begin{equation}
    W^{*} = \underset{W}{\arg\min} L_{train}(N(S, W)),
\end{equation}
where $S$ and $W$ denote the searching space and network parameters for the supernet $N(S, W)$, respectively. The $L_{train}$ means the training objective function. Then the optimal subnets $S^{*}$ are searched by the trained supernet as
\begin{equation}
    s^{*} = \underset{s \in S}{\arg\max} Acc_{val}(N(s, W^{*}(s))),
\end{equation}
where $Acc_{val}$ denotes the accuracy of the validation dataset. The final optimal network parameters $W^{'}$ is obtained by re-training the optimal subnet $M(\cdot)$ under data $X$ as  
\begin{equation}
    W^{'} = \underset{W}{\arg\min} L_{train}(M(X; W)).
\end{equation}

There is currently little discussion on the automation of UIE network design, and it deserves further exploration.

\begin{figure*}
    \small
    \centering
    \includegraphics[width=17.7cm,height=3.5cm]{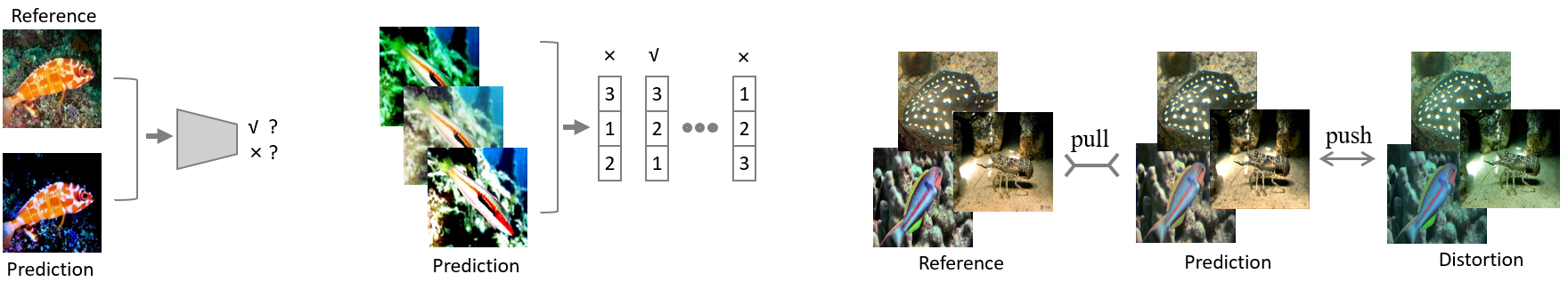}
    \leftline{\hspace{0.5cm} (a) Adversarial Learning \hspace{2cm} (b) Rank Learning \hspace{3.5cm} (c) Contrastive Learning}
    \caption{Schematic diagrams of the Adversarial Learning, Rank Learning and Contrastive Learning in the UIE task.}
    \label{fig:adversarial_contrastive_ranker}
  \end{figure*}

\subsection{Learning Strategy}
\label{subsec:learning_process}

\subsubsection{Adversarial Learning} 
The training strategy used in the generative adversarial network (GAN) \cite{goodfellow2020generative} can provide additional supervision signals for the image quality improvement process of the UIE task. The objective function of GAN is
\begin{equation}
    \mathcal{L}(G, D) = \mathbb{E}_{y \sim p(y)} \log{D(y)} + \mathbb{E}_{x \sim p(x)}[\log{(1 - D(G(x)))}],
\end{equation}
where $G$ and $D$ denote the generator and discriminator, respectively. For the UIE task, the generator $G$ is usually used to generate enhanced images or physical parameters. The process of distinguishing enhanced and reference underwater images through adversarial training is visualized in Fig. \ref{fig:adversarial_contrastive_ranker}-(a). An architecture with a dual generator and a single discriminator is designed by DGD-cGAN \cite{gonzalez2024dgd}. One generator is responsible for learning the mapping of distorted images to enhanced images, while another generator has the ability to simulate the imaging process through transmission information. In order to integrate information at different scales simultaneously, UIE-cGAN \cite{yang2020underwater} and TOPAL \cite{jiang2022target} adopt a single generator and dual discriminator architecture. One discriminator is responsible for optimizing local detailed features, while the other discriminator is designed to distinguish global semantic information. CE-CGAN \cite{agarwal2023contrast} and RUIG \cite{desai2021ruig} utilize the discriminative process provided by conditional generative adversarial networks as an auxiliary loss. By utilizing the adversarial training, enhanced images obtained by generator may be more consistent with the distribution of reference underwater high-quality images perceptually.

\subsubsection{Rank Learning}
The optimization goal of supervised loss is to make the enhanced image as close as possible to the reference image. However, model performance may be limited by imperfect reference images. Therefore, ranking learning is explored to provide ``better'' guidance to UIE models. From a ranking perspective, the UIE model does not need to know a priori which reference image is the best choice but looks for which reference image is the better choice. The process of selecting the best enhanced image through the ranking strategy is visualized in Fig. \ref{fig:adversarial_contrastive_ranker}-(b). URanker \cite{guo2023underwater} proposes that quality evaluation metrics designed for the UIE task can be used to guide the optimization of UIE models. To this end, a ranker loss is designed by URanker to rank the order of underwater images under the same scene content in terms of visual qualities. For $(x_{n}, x_{m})$ selected from the dataset $\{x_{0}, x_{1}, ..., x_{N}\}$ that contains $N$ images, URanker can predict the corresponding score $(s_{n}, s_{m})$ that conforms to the ranking relationship by 
\begin{equation}
    \mathcal{L}(s_{n}, s_{m}) = \begin{cases}
    \max{(0, (s_{m} -s_{n}) + \epsilon)}, q_{n} > q_{m}, \\
    \max{(0, (s_{n} -s_{m}) + \epsilon)}, q_{n} < q_{m}, \\
    \end{cases}
\end{equation}
where $\mathcal{L}(s_{n}, s_{m})$ is regarded as the margin-ranking loss. The $q_{n}$ and $q_{m}$ denote the quality of $x_{n}$ and $x_{m}$, respectively. Moreover, PDD-Net \cite{jiang2023perception} designs a pairwise quality ranking loss that is used in the form of a pre-trained model to guide the enhanced images toward the higher visual quality. The comparative mechanism is utilized by CLUIE-Net \cite{li2022beyond} to learn from multiple candidates of the reference. The main advantage of comparative learning is to make the image quality generated by the network better than the current enhancement candidates.
\begin{figure*}
    \small
    \centering
    \includegraphics[width=17.7cm,height=2.4cm]{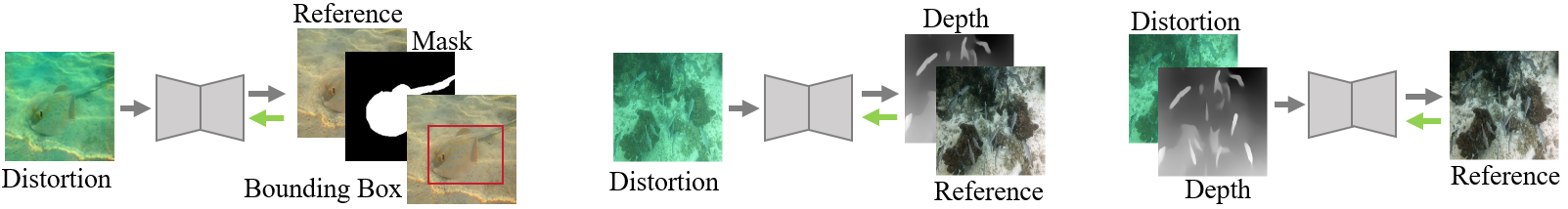}
    \leftline{\hspace{1cm} (a) Semantic Assistance  \hspace{3.5cm} (b) Guided by Depth \hspace{3cm} (c) Taking Depth as Input}
    \caption{Schematic diagrams of improving the UIE model performance by different auxiliary tasks. The gray and green arrows represent forward and backward propagation, respectively.}
    \label{fig:assistance_task}
  \end{figure*}

\subsubsection{Contrastive Learning}
Real-world pairs of distorted images and reference images are difficult to obtain. Therefore, reference images that are not accurate enough may provide imperfect supervision signals for the UIE network. Contrastive learning, a method that can alleviate the problem of missing labeled data, is used by UIE models to provide additional supervision signals. The optimization goals of UIE algorithms inspired by contrastive learning are generally consistent, which is to increase the distance between the anchor and the negative example while reducing the distance between the anchor and the positive example. The process of optimizing the distance between different types of examples through contrastive learning is visualized in Fig. \ref{fig:adversarial_contrastive_ranker}-(c). Research on contrast patterns in the UIE task focuses on two perspectives. The first is how to construct positive examples, negative examples and anchors. The second is how to optimize the distance between examples or features. TACL \cite{liu2022twin} treats the observed distorted underwater image as the negative example and the clear image as the positive example. To construct a reliable representation space, TACL adopts a pre-trained feature extractor to obtain the perceptual feature for the computing of the contrastive loss. A hybrid contrastive learning regularization $\mathcal{L}_{hclr}$ is proposed by HCLR-Net \cite{zhou2024hclr} as
\begin{equation}
    \mathcal{L}_{hclr} = \sum_{i=1}^{5} w_{i} \frac{D(G_{i}(y), G_{i}(\phi(x)))}{D(G_{i}(x_{r}, G_{i}(\phi(x))))},
\end{equation}
where $D(\cdot, \cdot)$ denotes the $\mathcal{L}_1$ loss. $G_{i}$ and $w_{i}$ represent the $i$-th layer of the pretrained feature extraction model and weight factor, respectively. $\phi(\cdot)$ is combined by feature and detail network branches. $x_{r}$ means a randomly selected distorted underwater image. The reliable bank and augmentation methods are used by Semi-UIR \cite{huang2023contrastive} to generate positive and negative examples. No-reference evaluation metrics are used in the triplet construction process of Semi-UIR as a guide for data selection. A closed-loop approach is adopted by TFUIE \cite{yu2023task}, which uses two paths to simultaneously enhance and synthesize distorted images. Correspondingly, a triplet contrastive loss is applied to both the enhancement and synthesis processes.

\subsubsection{Reinforcement Learning}
Deep learning has been widely proven to be effective for the UIE task. Beyond conventional research on neural networks that use fidelity losses for optimization, HPUIE-RL \cite{song2024hierarchical} explores how to take advantage of both deep learning and reinforcement learning. A two-stage pipeline with pre-training and fine-tuning is adopted by HPUIE-RL. The update method of model parameters during pre-training is consistent with that of conventional deep neural networks. For the fine-tuning phase of the model, a reinforcement learning strategy based on the reward function is designed. The reward function consists of three no-reference metrics that evaluate underwater image quality from different perspectives, which is 
\begin{equation}
    \mathcal{R} = \beta_{1} \times |\mathcal{R}_{a} - r^{max}_{a}| + \beta_{2} \times \mathcal{R}_{b} + \beta_{3} \times |\mathcal{R}_{c} - r^{max}_{c}|,
\end{equation}
where $a$, $b$ and $c$ represent the metric UCIQE \cite{yang2015underwater}, NIQE \cite{mittal2012making} and URanker \cite{guo2023underwater}, respectively. The $\beta_{1}$, $\beta_{2}$ and $\beta_{3}$ denote weight factors of the reward $\mathcal{R}_{a}$, $\mathcal{R}_{b}$ and $\mathcal{R}_{c}$, respectively. The $r^{max}_{a}$ and $r^{max}_{c}$ are the upper bounds to constrain the value range. The performance of HPUIE-RL can be continuously improved through iterative refinement optimization.

\subsection{Learning Stage}
\label{subsec:learning_stage}

\subsubsection{Single Stage} 
With the help of the data fitting capabilities of neural networks, the mapping from the distortion to the prediction can be learned in a single-stage manner as shown in Fig. \ref{fig:singleStage_twoStage_Diffusion}. Representative single-stage enhancement methods include UWCNN \cite{li2020underwater}, UWNet \cite{naik2021shallow} and UResNet \cite{liu2019underwater}, etc.

\subsubsection{Coarse-to-fine}
Considering the difficulty of obtaining optimal enhancement results at one time, the coarse-to-fine model with different purposes at different stages is utilized. By analyzing the color shifts and veil phenomena, a multi-branch multi-variable network for obtaining coarse results and attenuation factors is designed by MBANet \cite{xue2023investigating}. Then, the attenuation factors and coarse results are put into a physically inspired model to obtain refined enhancement results. A way to fuse two intermediate results through soft attention is proposed by GSL \cite{lin2022global}. One intermediate result contains structure and color information estimated by the global flow, while the other intermediate result handles overexposure and artifacts by the local flow. CUIE \cite{khandouzi2024coarse} builds a two-stage framework, in which the first stage obtains preliminary enhancement results through the global-local path, while the second stage uses histogram equalization and neural networks to improve the contrast and brightness of the image. The process of enhancing image quality by refinement in a two-stage manner proposed by CUIE is visualized in Fig. \ref{fig:singleStage_twoStage_Diffusion}-(b). A three-stage pipeline is designed by DMML \cite{esmaeilzehi2024dmml}, namely supervised training, adversarial training and fusion training. The goal of supervised training is to achieve high full-reference evaluation metric values, which usually represent better performance on synthetic data. Adversarial training is responsible for improving the no-reference evaluation metric values, which may make the model more friendly to real-world distorted images. The final fusion training takes advantage of both.

\subsubsection{Diffusion Process}
The diffusion model is a powerful generative model that has been widely studied recently. The forward process of the diffusion model is a Markov process that continuously adds noise to the image \cite{ho2020denoising}. High-quality images can be generated by the reverse process. The process of enhancing underwater image quality through the diffusion strategy is visualized in Fig. \ref{fig:singleStage_twoStage_Diffusion}-(c). The generative process cannot produce a specific enhanced underwater image in the desired form. Therefore, conditional information $c$ is added by UIE-DM \cite{tang2023underwater} to the diffusion process for the UIE task. The training loss of UIE-DM is
\begin{equation}
    \mathcal{L}_{s} = ||\epsilon_{t} - \epsilon_{\theta}(x_{t}, c, t)||_{1},
\end{equation}
where $x_{t}$ and $t$ denote the noisy image and time step, respectively. The $\epsilon_{t}$ and $\epsilon_{\theta}$ represent the noise image and predicted noise image, respectively. Therefore, the mean value $\mu_{\theta}$ can be defined as
\begin{equation}
    \mu_{\theta}(x_{t}, c, t) = \frac{1}{\sqrt{1 - \beta_{t}}} \left(x_{t} - \frac{\beta_{t}}{\sqrt{1 - \alpha_{t}}}\epsilon_{\theta}(x_{t}, c, t)\right).
\end{equation}

According to the above definition, the distorted underwater image can be used as condition information, which makes the image generated by the diffusion process deterministic. The relatively high computational cost of diffusion processes is an unsolved problem. SU-DDPM \cite{lu2023speed} proposes that the computational cost can be reduced by using different initial distributions. Although the diffusion model has been proven effective for the UIE task, its powerful data generation capabilities have not been fully explored.


\subsection{Assistance Task}
\label{subsec:assistance_task}

\subsubsection{Semantic Assistance}
High-level semantic information has not been widely explored in conventional UIE algorithms. In order to generate diverse features to UIE models, classification, segmentation and detection tasks have been embedded into UIE models by recent research. A schematic diagram of the UIE network training assisted by semantic information is visualized in Fig. \ref{fig:assistance_task}-(a). Regional information related to semantic segmentation maps is embedded into the UIE model by SGUIE \cite{qi2022sguie} to provide richer semantic information. A pre-trained classification model trained on a large-scale dataset is used as the feature extraction network by SATS \cite{wang2022semantic}. DAL-UIE \cite{kapoor2023domain} embeds a classifier that can constrain the latent space between the encoder and decoder. The classification loss $\mathcal{L}_{N}$ used for distinguish different water types is
\begin{equation}
    \mathcal{L}_{N} = -(1 - p_{t})^{\gamma} \log(p_{t}), 
\end{equation}
where $p_{t}$ is the probability of classification for category $t$ and $\gamma$ denotes the weight factor. Meanwhile, by minimizing the maximum mean divergence between the encoder and the classifier, robust features can be learned by DAL-UIE. A detection perception module is designed by WaterFlow \cite{zhang2023waterflow} to extract the local position information of the object. Through two-stage training of the UIE model and detection model from being independent to joint, the network can obtain features that represent position-related semantic information. Exploring how to combine low-level and high-level features can be beneficial for both tasks. The goal of the UIE task is not only to obtain images with good human subjective perception, but also to promote the performance of downstream tasks.

\begin{figure*}
    \small
    \centering
    \includegraphics[width=17.7cm,height=4.5cm]{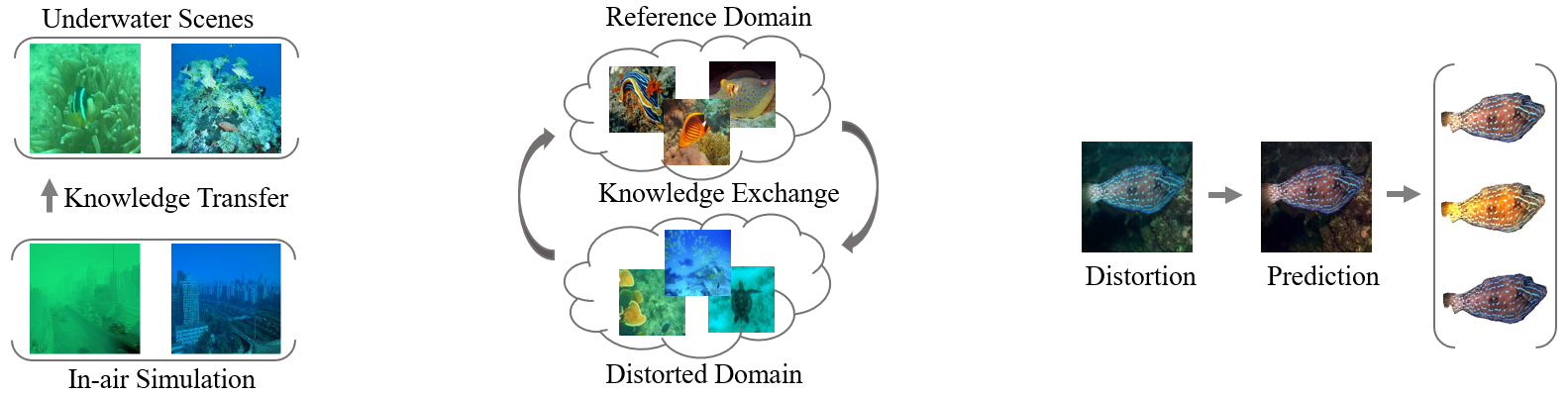}
    \leftline{\hspace{0.5cm} (a) Knowledge Transfer  \hspace{3cm} (b) Domain Translation \hspace{3cm} (c) Diversified Output}
    \caption{Schematic diagrams of UIE models designed from a domain perspective.}
    \label{fig:domain_perspective}
  \end{figure*}

\begin{figure*}
    \small
    \centering
    \includegraphics[width=17.7cm,height=4cm]{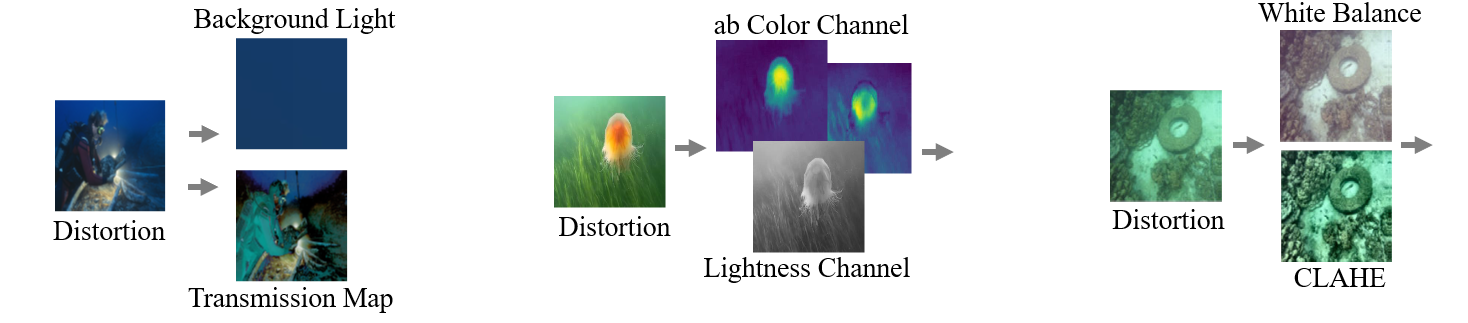}
    \leftline{\hspace{1cm} (a) Physical Embedding \hspace{3.5cm} (b) Color Space Fusion \hspace{3cm} (c) Multi-input Fusion}
    \caption{Schematic diagrams of disentanglement and fusion.}
    \label{fig:disentanglement_fusion}
  \end{figure*}

\subsubsection{Depth Assistance}
The degree of attenuation of underwater images is related to the depth of the scene. Therefore, depth maps are used by UIE algorithms to assist in the training of the network. As shown in Fig. \ref{fig:assistance_task}-(b) and Fig. \ref{fig:assistance_task}-(c), there are two ways to use depth information to assist the training of UIE models. The first one uses depth maps as the prediction targets, while the second one uses depth maps as the fusion inputs. Joint-ID \cite{yang2023joint} treats the UIE task and the depth estimation task as a unified task. Joint-ID uses a joint training method to enable the decoder to output enhanced images and depth images simultaneously. For the predicted depth map $\hat{d}$ and ground-truth  depth map $d$, the depth loss $\mathcal{L}_{depth}$ is  
\begin{equation}
    g_{j} = \log{\hat{d}_{j}} - \log{d_{j}},
\end{equation}
\begin{equation}
    \mathcal{L}_{depth}(\hat{d}, d) = \lambda_{1} \sqrt{\frac{1}{T} \sum_{j}{g_{j}^{2} - \frac{\lambda_{2}}{T^{2}}\left(\sum_{i}{g_{j}^{2}}\right)^{2}}}, 
\end{equation}
where $\lambda_{1}$ and $\lambda_{2}$ are two weight factors. The $j$ represents the index of the total number of pixels $T$. A combination of depth estimation loss $\mathcal{L}_{depth}$ and UIE loss is used for the training of Joint-ID. A two-stage architecture is designed by DAUT \cite{badran2023daut}, whose first and second stages are depth estimation and enhancement, respectively. In the enhancement stage, namely the second stage, the depth map and the distortion map are simultaneously used as inputs to the enhancement network to provide richer prior information. DepthCue \cite{desai2023depthcue} utilizes the depth map obtained by a pre-trained depth estimation network as auxiliary information for the decoder of the enhancement network. HybrUR \cite{yan2023hybrur} trains a depth estimation network from scratch that provides depth information which fits the degeneration factors. Although these studies show that depth maps can bring positive effects to the UIE task, it is worth pointing out that the monocular depth estimation task itself has inherent experimental errors, since it is almost impossible to obtain perfect depth information from a monocular image.

\subsection{Domain Perspective}
\label{subsec:domain_perspective}

\subsubsection{Knowledge Transfer}
Synthetic data cannot accurately reflect the complex underwater attenuation of the real world. Therefore, a model trained on synthetic data may not achieve satisfactory performance in the real-world UIE task. Domain discrepancy between synthetic and real-world data are inherent. Therefore, domain adaptation strategies are explored to reduce this domain bias. A triple-alignment network with translation path and a task-oriented enhancement path is designed by TUDA \cite{wang2023domain}.  Domain gap is handled at image-level ($\mathcal{L}^{img}$), output-level ($\mathcal{L}^{out}$), feature-level ($\mathcal{L}^{feat}$) by discriminators $D^{img}$, $D^{out}$ and $D^{feat}$, respectively. The inter-domain adaptation at image-level and output-level involves the real-world underwater image $x_{r}$ and the corresponding enhanced version $y_{r}$ as
\begin{equation}
\begin{split}
    \mathcal{L}^{img} & = \mathbb{E}_{x_{st}}[D^{img}(x_{st})] - \mathbb{E}_{x_{r}}[D^{img}(x_{r})]  \\
            &= + \lambda_{img} \mathbb{E}_{\hat{I}}(||\nabla_{\hat{I}}D^{img}(\hat{I})||_{2} - 1)^{2},
\end{split}
\end{equation}
\begin{equation}
\begin{split}
    \mathcal{L}^{out} & = \mathbb{E}_{y_{st}}[D^{out}(y_{st})] - \mathbb{E}_{y_{r}}[D^{out}(y_{r})] \\
            & + \lambda_{out} \mathbb{E}_{\hat{I}}(||\nabla_{\hat{I}}D^{out}(\hat{I})||_{2} - 1)^{2},
\end{split}
\end{equation}
where $x_{st}$ is translated from an in-air image $x_{s}$, while the $y_{st}$ is an enhanced version of $x_{st}$. The $\hat{I}$ is sampled from $\{x_{r}, x_{st}\}$. The domain adaptation at feature-level is
\begin{equation}
    \begin{split}
        \mathcal{L}^{feat} & = \mathbb{E}_{x_{st}}[D^{feat}(\mathcal{G}_{r}(x_{st}))] - \mathbb{E}_{x_{r}}[D^{feat}(\mathcal{G}_{r}(x_{r}))] \\
                 & + \lambda_{feat} \mathbb{E}_{\hat{I}}(||\nabla_{\hat{I}}D^{feat}(\hat{I})||_{2} - 1)^{2},
    \end{split}
\end{equation}
where $\mathcal{G}_{r}$ is the inter-class encoder. $\lambda_{img}$, $\lambda_{out}$ and $\lambda_{feat}$ are weight factors. The schematic diagram of knowledge transfer is shown in Fig. \ref{fig:domain_perspective}-(a). By utilizing domain transformation algorithms, in-air images and underwater images are simultaneously used by TSDA \cite{jiang2022two} to construct intermediate domains. Then, the domain discrepancy between the intermediate domain and in-air domain is reduced by a carefully designed enhancement network. Unpaired underwater images and paired in-air images are used by IUIE \cite{bing2023domain} for semi-supervised training. Through weight sharing strategy and channel prior loss, in-air images and prior knowledge are used to enhance the quality of underwater distorted images. WSDS-GAN \cite{liu2023wsds} designs a weak-strong two-stage process, in which weak learning and strong learning are implemented in unsupervised and supervised ways, respectively. Weak learning uses in-air and underwater image domains to learn information such as the content and brightness, while the goal of strong learning is to reduce the blurring of details caused by training with domain differences.

\subsubsection{Domain Translation}
Distorted and clear underwater images can be viewed as two different domains. Through unsupervised domain transformation, the dependence on pairs of underwater images can be alleviated. The unsupervised disentanglement of content information and style information is designed by URD-UIE \cite{zhu2023unsupervised}. URD-UIE proposes that in the UIE task, content information usually represents the texture and semantics, while style information can represent degradation such as noise or blur. By leveraging an adversarial process to learn content and style encoding, the degradation information can be manipulated in the latent space. Aiming at achieving cross-domain translation, URD-UIE employs a domain bidirectional loop reconstruction process, that is, $x \rightarrow x_{x-y} \rightarrow \dot{x}$ and $y \rightarrow y_{y-x} \rightarrow \dot{y}$, where $\dot{x}$ and $\dot{y}$ represent reconstructed image after two-time domain translations. The reconstruction is adopted on the image ($\mathcal{L}^{img}$), style ($\mathcal{L}^{sty}$) and content ($\mathcal{L}^{con}$) spaces as 
\begin{equation}
    \begin{cases}
    \mathcal{L}^{img} = ||x - \dot{x}||_{1} + ||y - \dot{y}||_{1}, \\
    \mathcal{L}^{sty} = ||s_x - \dot{s_{x}}||_{1} + ||s_y - \dot{s_{y}}||_{1}, \\
    \mathcal{L}^{con} = ||c_x - \dot{c_{x}}||_{1} + ||c_y - \dot{c_{y}}||_{1}, \\
    \end{cases}
\end{equation}
where $\{s_{x}, s_{y}\}$ and $\{c_{x}, c_{y}\}$ denote style and content codes for $x$ and $y$, respectively. The $\dot{s_{x}}, \dot{s_{y}}, \dot{c_{x}}$ and $\dot{c_{y}}$ mean the corresponding reconstructed versions. TACL \cite{liu2022twin} constructs a closed-loop path to learn bidirectional mappings simultaneously. The forward branch is responsible for learning distortion to clean images, while the optimization goal of the reverse network is to learn the attenuation process. The process of domain translation is shown in Fig. \ref{fig:domain_perspective}-(b).

\subsubsection{Diversified Output}
For an image enhancement task which has no perfect solution, it may be beneficial to provide the user with selectable and diverse outputs. The pixel-wise wasserstein autoencoder architecture is designed by PWAE \cite{kim2021pixel}, which has a two-dimensional latent tensor representation. The tensor $z_{h}$ and $z_{s}$ are used to represent enhancement space and style space. The space translation can be implemented by  
\begin{equation}
    z_{h \rightarrow s} = \sigma(z_{s})\left(\frac{z_{h} - \mu{(z_{h})}}{\sigma(z_{h})}\right) + \mu{(z_{s})},
\end{equation}
where $\mu(\cdot)$ and $\sigma(\cdot)$ denote the mean and standard deviation, respectively. The degree of the fused tensor $z^{s}_{h}$ can be controlled by the parameter $\alpha$ as
\begin{equation}
    z^{s}_{h} = \alpha z_{h \rightarrow s} + (1 - \alpha)z_{h}.
\end{equation}

By fusing the latent code provided by the style image, PWAE is able to change the illumination and style of the enhanced result. Inspired by the multi-domain image-to-image algorithm, UIESS \cite{chen2022domain} decomposes the UIE process into content and style learning flows. By manipulating the encoding values in the latent space, images with different degrees of enhancement can be produced by the decoder of UIESS. Beyond the content and style codes studied by UIESS, CECF \cite{cong2024underwater} proposes the concept of color code. CECF assumes that when there exists images with less local distortion in the dataset, these images may have colors with long wavelengths that are approximately invariant during the enhancement process. By learning a color representation with such invariance, CECF can obtain color-adjustable underwater creatures from guidance images. The diverse output provided by CECF is shown in Fig. \ref{fig:domain_perspective}-(c).


\begin{table*}
    
    \centering
    \scriptsize
    \setlength{\tabcolsep}{2.9mm}
    \renewcommand{\arraystretch}{1.4}
      \flushleft
      \caption{Quantitative results with supervised training on full-reference benchmarks in terms of PSNR and SSIM. Best results are in \textcolor{red}{red} and second best results are in \textcolor{blue}{blue}.}
      \label{tab:full_reference_results}
        
        \begin{tabular}{@{}c|cc|cc|cc|cc|cc|cc@{}}
          \toprule
          \multirow{2}{*}{Method} & \multicolumn{2}{c|}{UIEB} & \multicolumn{2}{c|}{LSUI} & \multicolumn{2}{c|}{EUVP-D} & \multicolumn{2}{c|}{EUVP-I} & \multicolumn{2}{c|}{EUVP-S} & \multicolumn{2}{c}{UFO-120} \\
          
          & PSNR$\uparrow$ & SSIM$\uparrow$ & PSNR$\uparrow$ & SSIM$\uparrow$ & PSNR$\uparrow$ & SSIM$\uparrow$ & PSNR$\uparrow$ & SSIM$\uparrow$ & PSNR$\uparrow$ & SSIM$\uparrow$ & PSNR$\uparrow$ & SSIM$\uparrow$  \\
          \midrule
          
          UWNet \cite{naik2021shallow}      & 17.36 & 0.7948 & 22.01 & 0.8489 & 21.36 & 0.8887 & 22.54 & 0.8115 & 25.45 & 0.8616 & 25.02 & 0.8505 \\
          PhysicalNN \cite{chen2021underwater}   & 18.08 & 0.7693 & 22.52 & 0.8468 & 21.37 & 0.8851 & 23.68 & 0.8108 & 26.07 & 0.8507 & 25.68 & 0.8414 \\
          FUnIEGAN \cite{islam2020fast}    & 19.12 & 0.8321 & 24.23 & 0.8579 & 21.04 & 0.8471 & 23.06 & 0.7731 & 27.55 & 0.8695 & 26.79 & 0.8378 \\
          WaterNet \cite{li2019underwater}    & 21.04 & 0.8601 & 22.74 & 0.8560 & 21.40 & 0.8881 & 23.99 & 0.8325 & 23.43 & 0.8287 & 23.84 & 0.8188 \\
          ADMNNet \cite{yan2022attention}     & 20.77 & 0.8709 & 26.11 & 0.9051 & 22.29 & 0.9063 & 24.65 & 0.8424 & 27.11 & 0.8893 & 26.67 & 0.8757 \\
          UColor \cite{li2021underwater}      & 20.13 & 0.8769 & 24.03 & 0.8855 & \textcolor{blue}{22.43} & \textcolor{red}{0.9088} & 24.43 & 0.8462 & 27.65 & \textcolor{red}{0.9036} & 26.95 & \textcolor{red}{0.8877} \\
          UGAN \cite{fabbri2018enhancing}        & 21.37 & 0.8534 & 26.93 & 0.8983 & 22.02 & 0.8850 & 24.71 & 0.8276 & 28.26 & 0.8851 & 27.61 & 0.8672 \\
          SGUIE \cite{qi2022sguie}       & 21.12 & 0.8882 & 27.77 & 0.9129 & 22.06 & 0.8927 & 24.68 & 0.8356 & 28.83 & 0.8934 & 28.30 & 0.8784 \\
          UIE-WD \cite{ma2022wavelet}      & 20.10 & 0.8617 & 26.81 & 0.9088 & 22.21 & 0.9080 & 24.56 & 0.8365 & 28.43 & \textcolor{blue}{0.8990} & 27.48 & 0.8756 \\
          SCNet \cite{fu2022underwater}       & 21.37 & 0.8981 & 27.47 & 0.9111 & 22.40 & 0.9028 & 24.69 & 0.8267 & 28.29 & 0.8850 & 27.34 & 0.8690 \\
          STSC \cite{wang2022semantic}        & 21.05 & 0.8661 & 27.67 & 0.9108 & \textcolor{red}{22.45} & \textcolor{blue}{0.9084} & \textcolor{red}{25.27} & 0.8484 & \textcolor{blue}{29.05} & 0.8956 & 28.16 & \textcolor{blue}{0.8847} \\
          U-Trans \cite{peng2023u}     & 20.39 & 0.8034 & 27.48 & 0.9013 & 22.15 & 0.8941 & 24.56 & 0.8309 & 28.46 & 0.8843 & 27.60 & 0.8525 \\
          CECF \cite{cong2024underwater}        & 21.82 & 0.8866 & 27.27 & 0.9124 & 22.21 & 0.9032 & 24.86 & \textcolor{red}{0.8567} & 27.13 & 0.8879 & 26.75 & 0.8729 \\
          Semi-UIR \cite{huang2023contrastive}    & \textcolor{blue}{22.79} & \textcolor{blue}{0.9088} & 25.44 & 0.8801 & 22.31 & 0.9059 & 24.46 & 0.8170 & 26.07 & 0.8402 & \textcolor{blue}{28.41} & 0.8837 \\
          UIE-DM \cite{tang2023underwater}      & \textcolor{red}{23.03} & \textcolor{red}{0.9099} & \textcolor{red}{29.56} & \textcolor{red}{0.9194} & 22.13 & 0.9066 & \textcolor{blue}{25.14} & \textcolor{blue}{0.8499} & 26.12 & 0.8757 & 26.10 & 0.8464 \\
          HCLR-Net \cite{zhou2024hclr}    & 22.09 & 0.8976 & \textcolor{blue}{27.94} & \textcolor{blue}{0.9169} & \textcolor{blue}{22.43} & 0.9075 & 24.92 & 0.8314 & \textcolor{red}{29.23} & 0.8928 & \textcolor{red}{28.85} & 0.8818 \\
          GUPDM \cite{mu2023generalized}       & 21.76 & 0.8697 & 27.34 & 0.9116 & 22.22 & 0.8739 & 24.91 & 0.8330 & 28.26 & 0.8904 & 27.81 & 0.8828 \\
          \bottomrule
        
        \end{tabular}
      
\end{table*}

\begin{figure*}
    \centering
    
    \includegraphics[width=2.8cm,height=2.2cm]{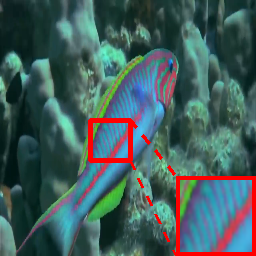}
    \includegraphics[width=2.8cm,height=2.2cm]{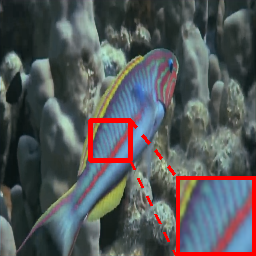}
    \includegraphics[width=2.8cm,height=2.2cm]{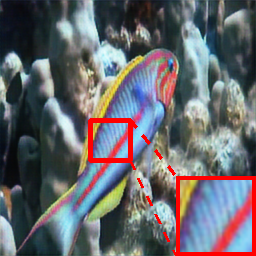}
    \includegraphics[width=2.8cm,height=2.2cm]{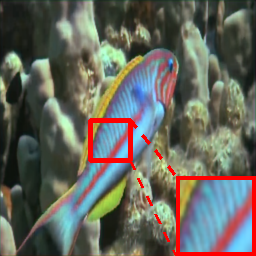}
    \includegraphics[width=2.8cm,height=2.2cm]{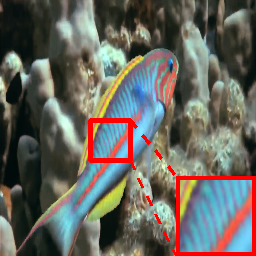}
    \includegraphics[width=2.8cm,height=2.2cm]{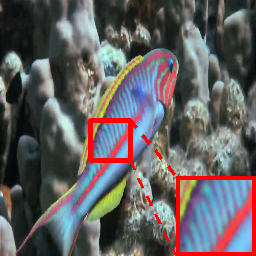}

    \includegraphics[width=2.8cm,height=2.2cm]{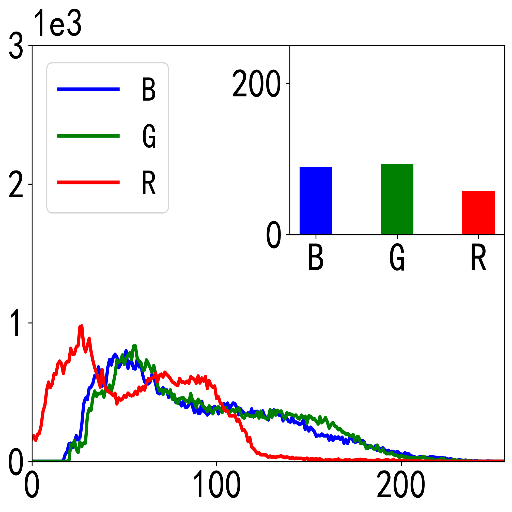}
    \includegraphics[width=2.8cm,height=2.2cm]{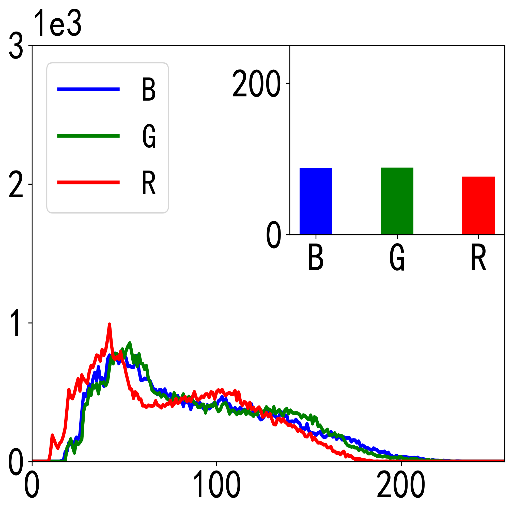}
    \includegraphics[width=2.8cm,height=2.2cm]{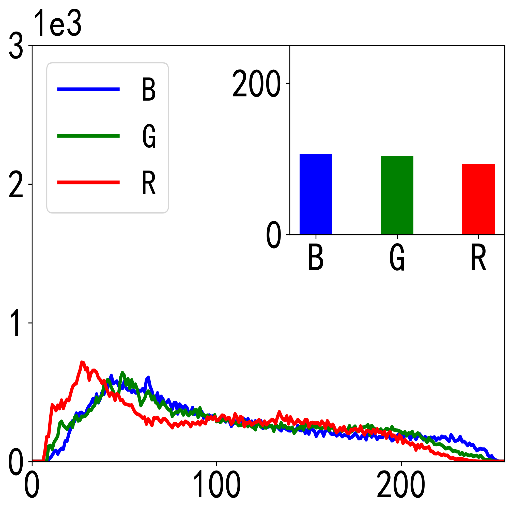}
    \includegraphics[width=2.8cm,height=2.2cm]{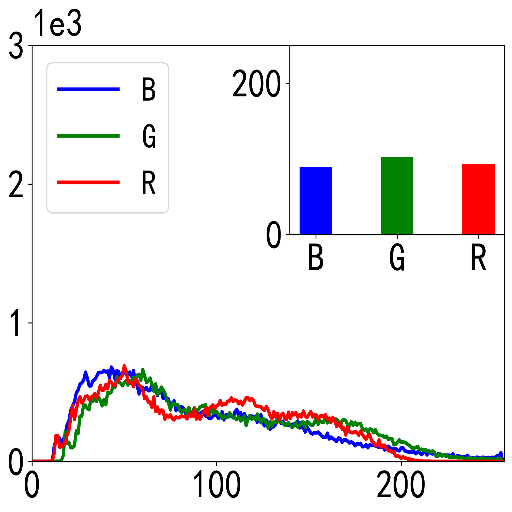}
    \includegraphics[width=2.8cm,height=2.2cm]{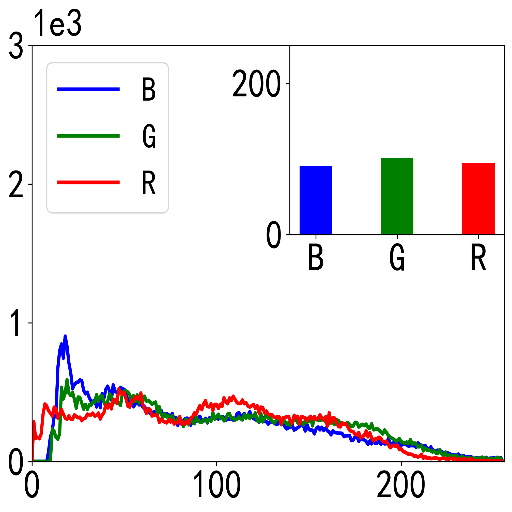}
    \includegraphics[width=2.8cm,height=2.2cm]{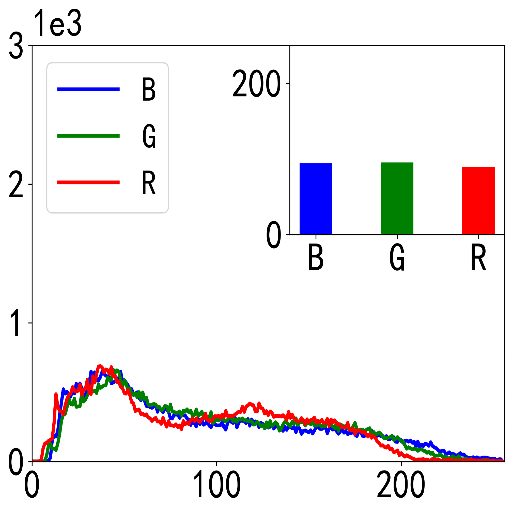}

    \leftline{\hspace{1cm} Distortion \hspace{1.3cm} UWNet \hspace{1.4cm} FUnIEGAN \hspace{1.1cm} WaterNet \hspace{1.2cm} ADMNNet \hspace{1.2cm} UColor}

    \vspace{0.1cm}

    \includegraphics[width=2.8cm,height=2.2cm]{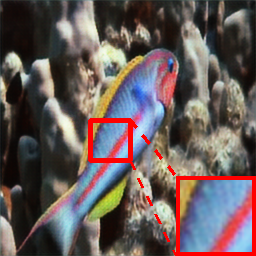}
    \includegraphics[width=2.8cm,height=2.2cm]{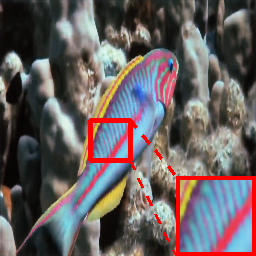}
    \includegraphics[width=2.8cm,height=2.2cm]{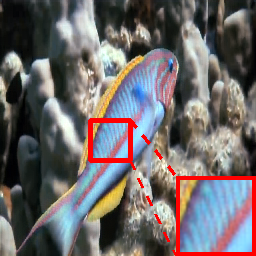}
    \includegraphics[width=2.8cm,height=2.2cm]{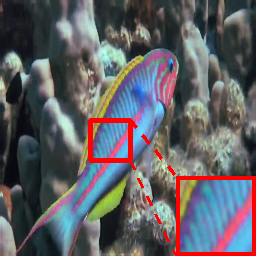}
    \includegraphics[width=2.8cm,height=2.2cm]{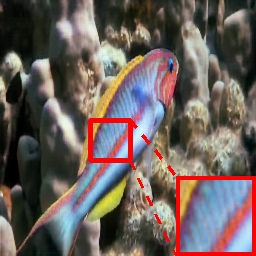}
    \includegraphics[width=2.8cm,height=2.2cm]{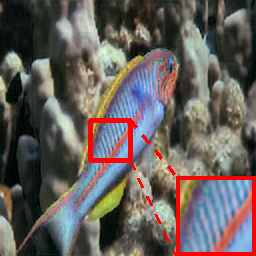}

    \includegraphics[width=2.8cm,height=2.2cm]{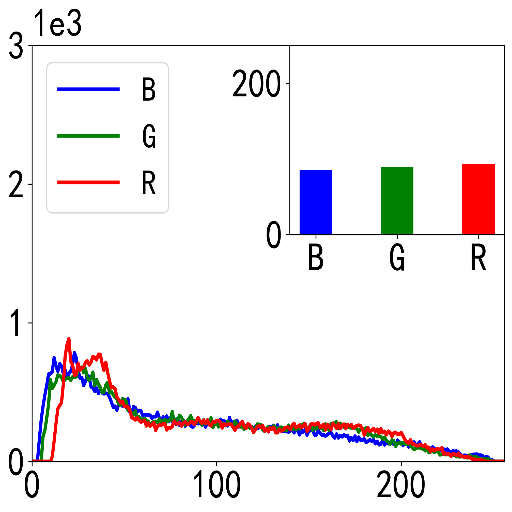}
    \includegraphics[width=2.8cm,height=2.2cm]{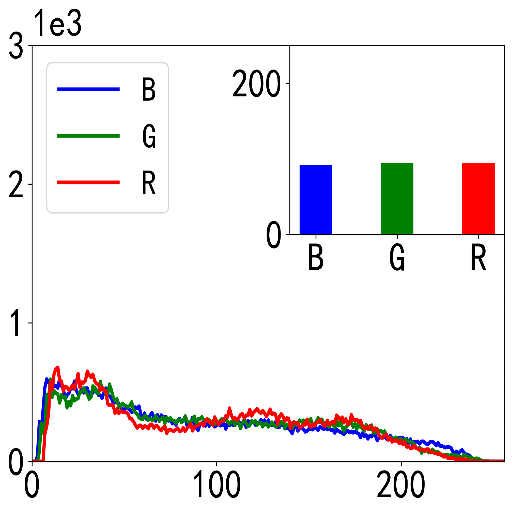}
    \includegraphics[width=2.8cm,height=2.2cm]{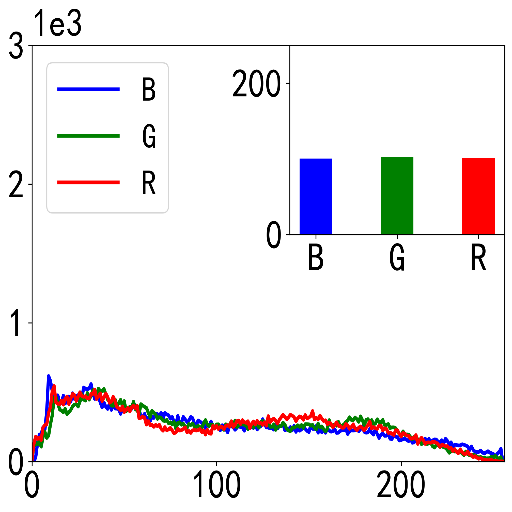}
    \includegraphics[width=2.8cm,height=2.2cm]{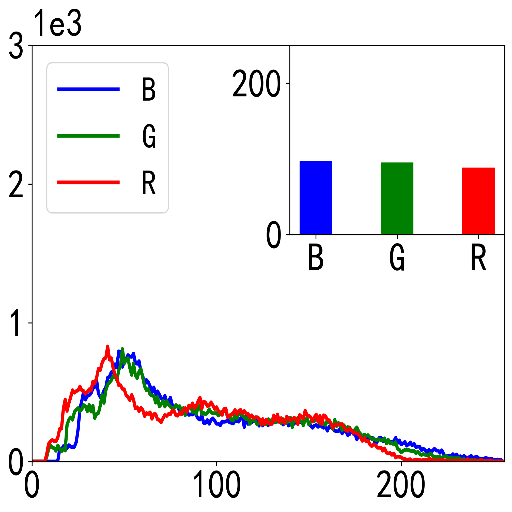}
    \includegraphics[width=2.8cm,height=2.2cm]{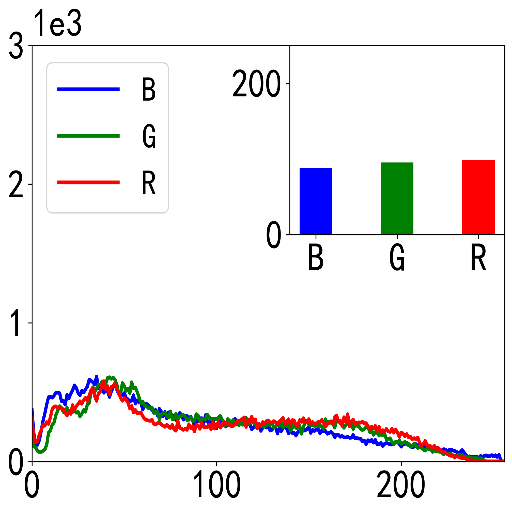}
    \includegraphics[width=2.8cm,height=2.2cm]{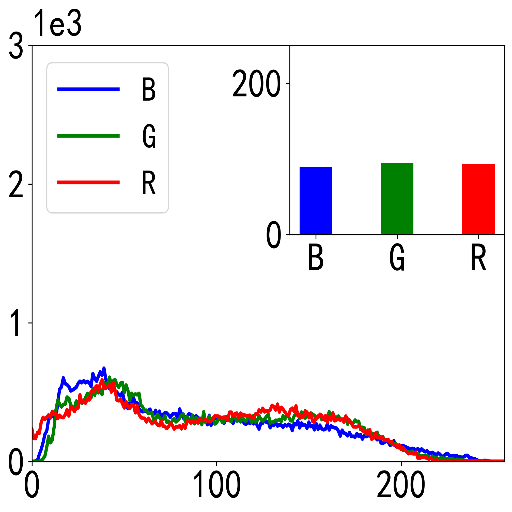}

    \leftline{\hspace{1.2cm} UGAN \hspace{1.4cm} SGUIE \hspace{1.7cm} UIE-WD \hspace{1.5cm} SCNet \hspace{1.7cm} STSC \hspace{1.8cm} U-Trans}
    
    \vspace{0.1cm}

    \includegraphics[width=2.8cm,height=2.2cm]{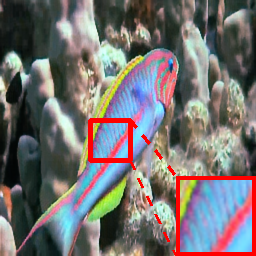}
    \includegraphics[width=2.8cm,height=2.2cm]{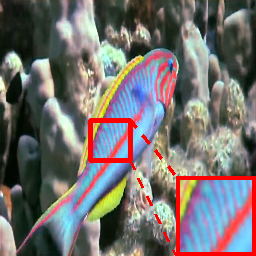}
    \includegraphics[width=2.8cm,height=2.2cm]{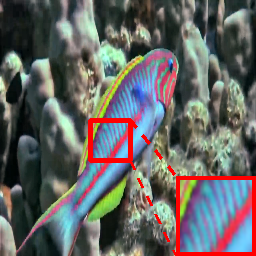}
    \includegraphics[width=2.8cm,height=2.2cm]{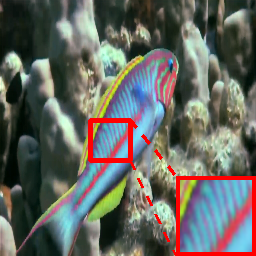}
    \includegraphics[width=2.8cm,height=2.2cm]{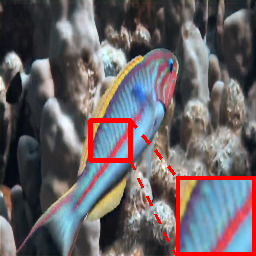}
    \includegraphics[width=2.8cm,height=2.2cm]{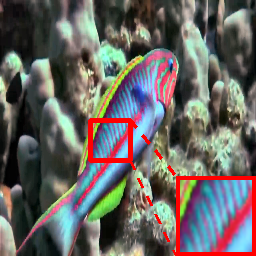}

    \includegraphics[width=2.8cm,height=2.2cm]{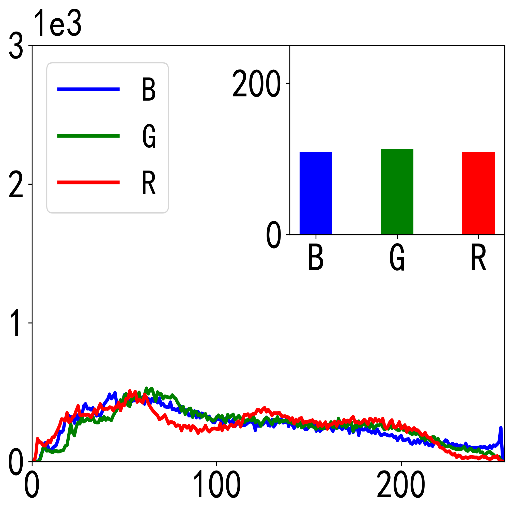}
    \includegraphics[width=2.8cm,height=2.2cm]{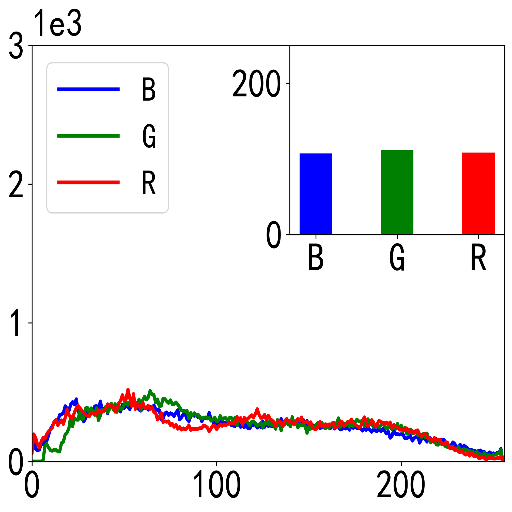}
    \includegraphics[width=2.8cm,height=2.2cm]{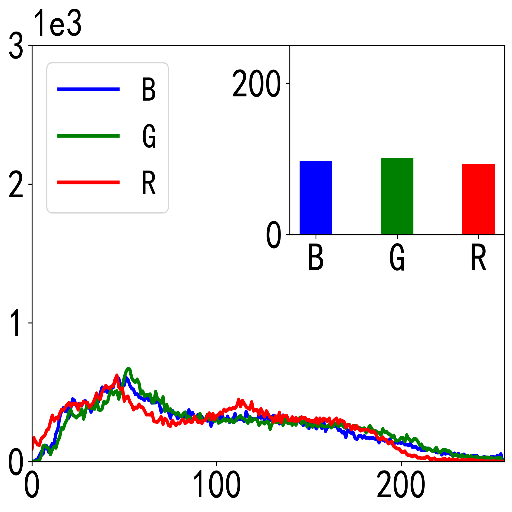}
    \includegraphics[width=2.8cm,height=2.2cm]{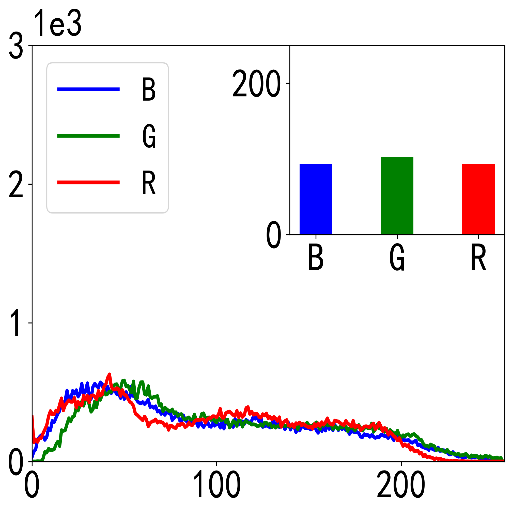}
    \includegraphics[width=2.8cm,height=2.2cm]{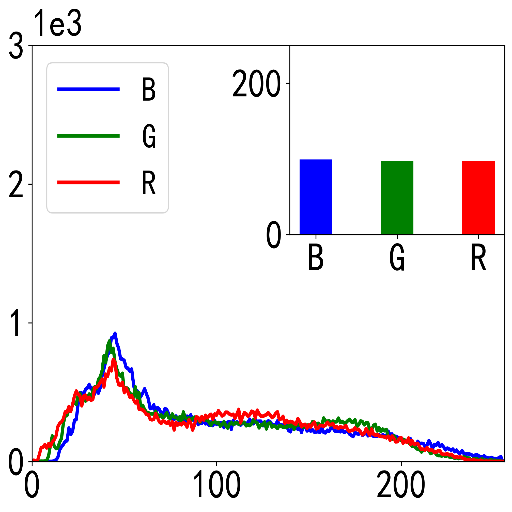}
    \includegraphics[width=2.8cm,height=2.2cm]{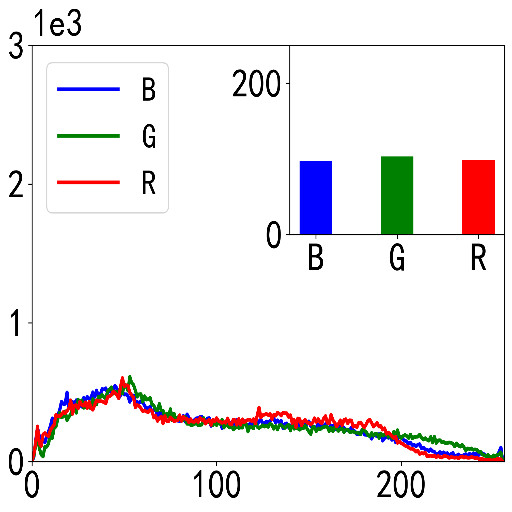}

    \leftline{\hspace{1.3cm} CECF \hspace{1.3cm} Semi-UIR \hspace{1.4cm} UIE-DM \hspace{1.3cm} HCLR-Net \hspace{1.2cm} GUPDM \hspace{1.3cm} Reference}

    \caption{Visual results obtained on UIEB full-reference test set. The curve figure denotes the pixel histogram. The bar figure gives the average pixel value of each channel.}
    \label{fig:visual_results_uieb_2774}
  \end{figure*}

\begin{table}
    \centering
    \scriptsize
    \setlength{\tabcolsep}{0.2mm}
    \renewcommand{\arraystretch}{1.4}
    \flushleft
    \caption{Quantitative results on no-reference benchmarks in terms of UIQM and UCIQE. Best results are in \textcolor{red}{red} and second best results are in \textcolor{blue}{blue}.}
    \label{tab:non_reference_results}
    \begin{tabular}{@{}c|cc|cc|cc|cc@{}}
       
      \toprule
      \multirow{2}{*}{\textbf{Method}} & \multicolumn{2}{c|}{\textbf{Challenge-60}} & \multicolumn{2}{c|}{\textbf{U45}} & \multicolumn{2}{c|}{\textbf{UCCS}} & \multicolumn{2}{c}{\textbf{EUVP-330}}\\
      & \textbf{UIQM}$\uparrow$  & \textbf{UCIQE}$\uparrow$  & \textbf{UIQM}$\uparrow$  & \textbf{UCIQE}$\uparrow$  & \textbf{UIQM}$\uparrow$  & \textbf{UCIQE}$\uparrow$  &\textbf{UIQM}$\uparrow$  & \textbf{UCIQE}$\uparrow$ \\
      \midrule
      Distortion    & 3.041 & 0.5187  & 2.347 & 0.5306  & 1.698 & 0.4949  & 2.264 & 0.5268 \\

      \midrule
      {UWNet}       & 3.832 & 0.5061     & 3.792 & 0.5183      & 3.194 & 0.4821      & 3.429 & 0.5160   \\
      {PhysicalNN}   & 3.946 & 0.5444     & 3.797 & 0.5578      & 3.317 & 0.5067      & 3.368 & 0.5475   \\
      {FUnIEGAN}    & \textcolor{blue}{4.325} & 0.5819     & 4.372 & \textcolor{red}{0.6174}      & 4.086 & 0.5481      & 3.636 & 0.5964   \\
      {WaterNet}    & 4.195 & \textcolor{blue}{0.5850}     & 4.167 & 0.5960      & 4.158 & 0.5654      & 3.810 & \textcolor{blue}{0.6026}   \\
      {ADMNNet}     & 4.190 & 0.5649     & 4.277 & 0.5836      & 4.062 & 0.5477      & 3.748 & 0.5775   \\
      {UColor}      & 3.992 & 0.5565     & 4.144 & 0.5793      & 3.583 & 0.5357      & 3.552 & 0.5694   \\
      {UGAN}        & \textcolor{red}{4.480} & \textcolor{red}{0.5891}     & \textcolor{red}{4.501} & 0.6014      & \textcolor{red}{4.391} & 0.5654      & \textcolor{blue}{3.899} & \textcolor{red}{0.6153}   \\
      {SGUIE}       & 4.221 & 0.5791     & 4.256 & 0.6000      & 3.967 & 0.5525      & 3.663 & 0.5927   \\
      {UIE-WD}      & 4.102 & 0.5654     & 4.164 & 0.6004      & 3.711 & 0.5264      & 3.557 & 0.5810   \\
      {SCNet}       & 4.259 & 0.5813     & \textcolor{blue}{4.456} & 0.6005      & \textcolor{blue}{4.386} & \textcolor{red}{0.5727}      & 3.857 & 0.5991  \\
      {STSC}        & 4.261 & 0.5751     & 4.343 & 0.5988      & 4.004 & 0.5516      & 3.717 & 0.5923  \\
      {U-Trans}     & 4.063 & 0.5604     & 4.120 & 0.5914      & 3.819 & 0.5365      & 3.525 & 0.5746   \\
      {CECF}        & 4.127 & 0.5728     & 4.354 & 0.5918      & 4.286 & 0.5531      & 3.681 & 0.5903  \\
      {Semi-UIR}    & 4.098 & 0.5761     & 4.431 & 0.5999      & 4.049 & 0.5499      & 3.660 & 0.5974   \\
      {UIE-DM}      & 3.874 & 0.5639     & 4.097 & 0.5974      & 4.166 & \textcolor{blue}{0.5673}      & 3.652 & 0.5984   \\
      {HCLR-Net}    & 4.100 & 0.5656     & 4.293 & \textcolor{blue}{0.6027}      & 3.935 & 0.5376      & 3.680 & 0.5861   \\
      {GUPDM}       & 4.276 & 0.5652     & 4.399 & 0.5897      & 4.108 & 0.5300      & \textcolor{red}{4.017} & 0.5823   \\
      \bottomrule
    \end{tabular}
  
  \end{table}

\begin{figure*}
    \small
    \centering
    \includegraphics[width=8.5cm,height=4cm]{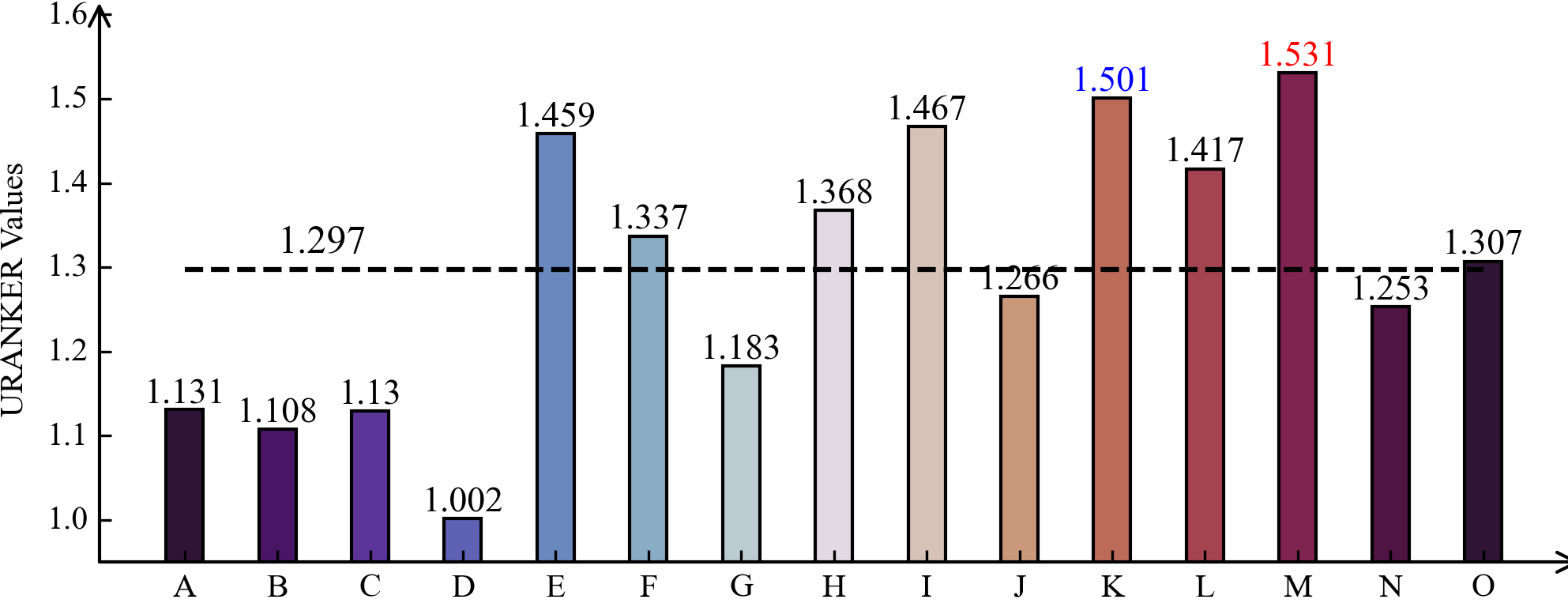}
    \includegraphics[width=8.5cm,height=4cm]{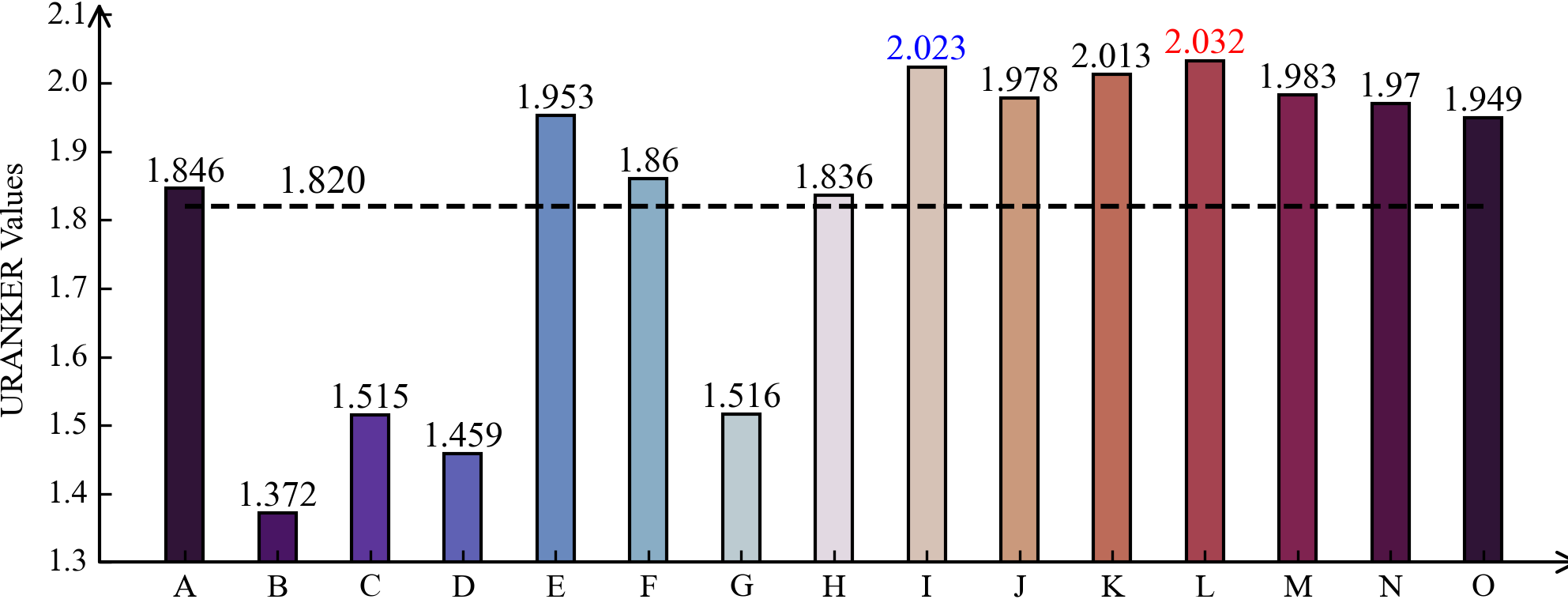}
    \leftline{\hspace{4cm} (a) Challenge-60 \hspace{6.5cm} (b) U45}

    \includegraphics[width=8.5cm,height=4cm]{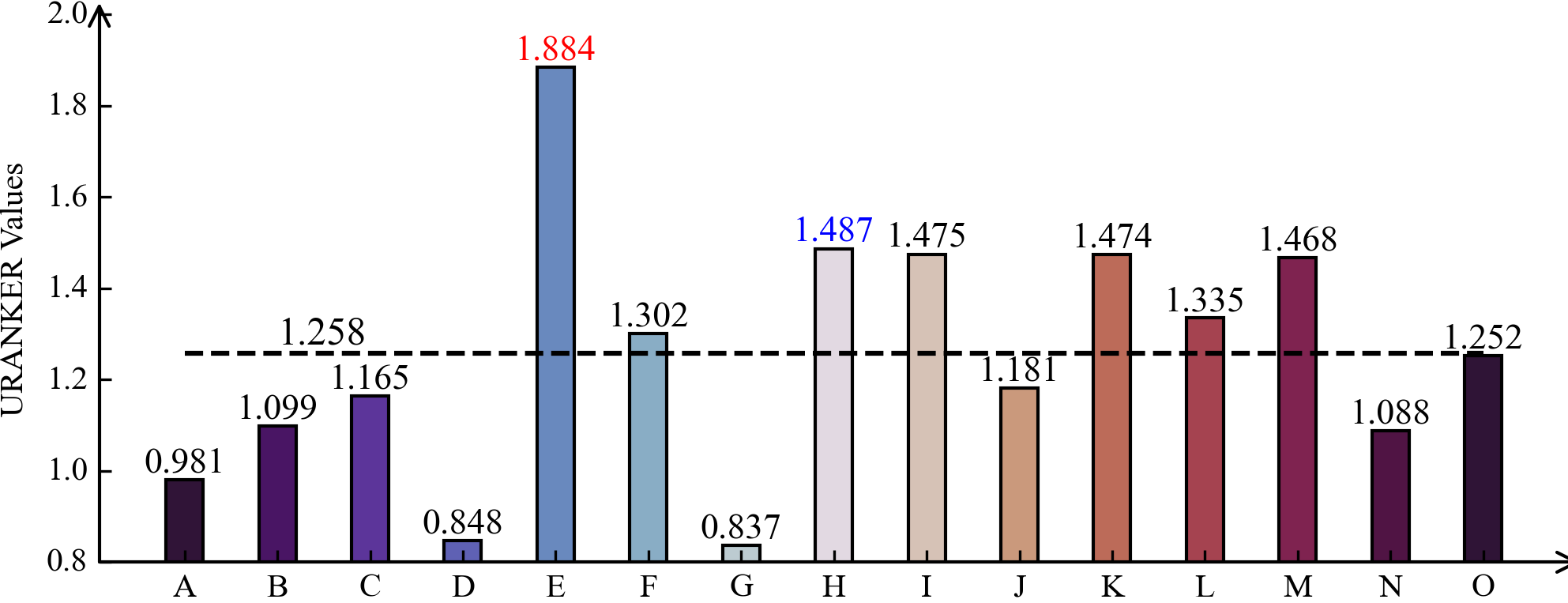}
    \includegraphics[width=8.5cm,height=4cm]{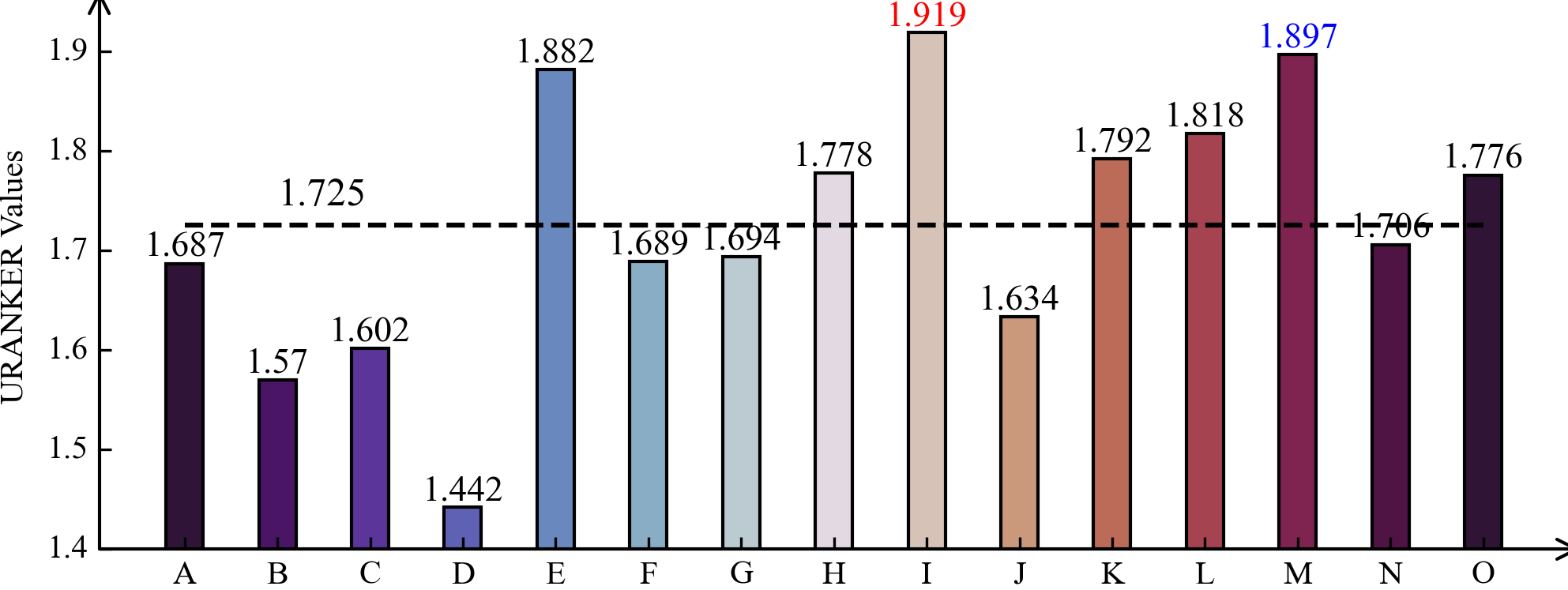}
    \leftline{\hspace{4cm} (c) UCCS \hspace{7cm} (d) EUVP-330}
    \caption{URANKER values on no-reference benchmark datasets. The horizontal dashed line represents the mean of the quantitative results obtained by all algorithms. The correspondence between the letters on the horizontal axis and the algorithm is: A (FUnIEGAN), B (WaterNet), C (ADMNNet), D (UColor), E (UGAN), F (SGUIE), G (UIE-WD),  H (SCNet), I (STSC), J (U-Trans), K (CECF), L (Semi-UIR), M (UIE-DM), N (HCLR-Net) and O (GUPDM). Best results are in \textcolor{red}{red font} and second best results are in \textcolor{blue}{blue font}.}
    \label{fig:nriqa}
  \end{figure*}

\begin{figure*}
    \centering
    \includegraphics[width=2.8cm,height=2.2cm]{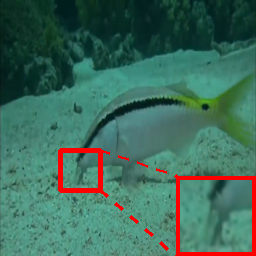}
    \includegraphics[width=2.8cm,height=2.2cm]{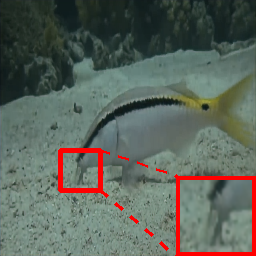}
    \includegraphics[width=2.8cm,height=2.2cm]{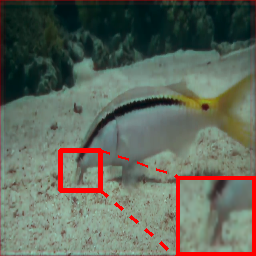}
    \includegraphics[width=2.8cm,height=2.2cm]{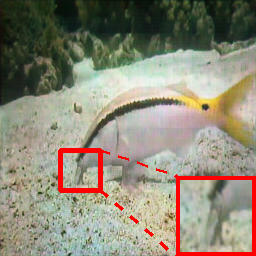}
    \includegraphics[width=2.8cm,height=2.2cm]{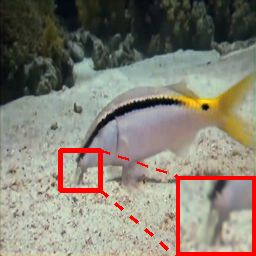}
    \includegraphics[width=2.8cm,height=2.2cm]{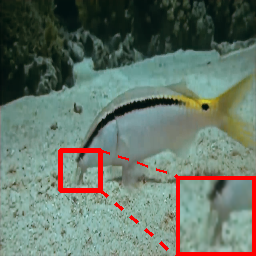}

    \includegraphics[width=2.8cm,height=2.2cm]{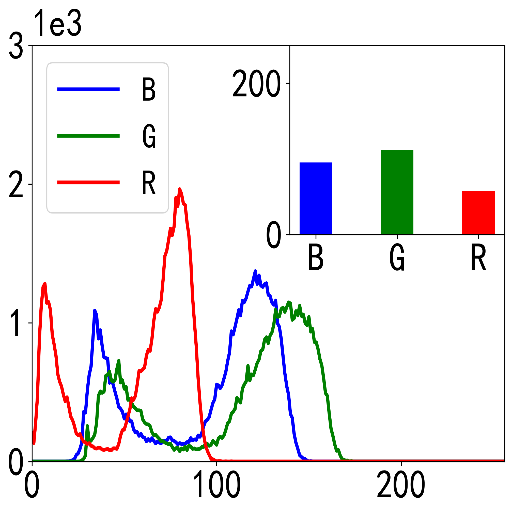}
    \includegraphics[width=2.8cm,height=2.2cm]{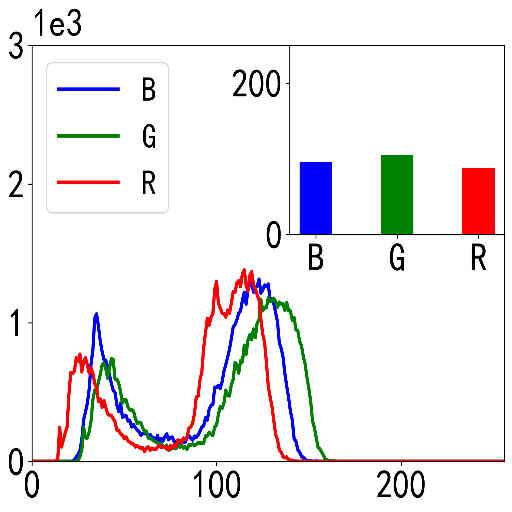}
    \includegraphics[width=2.8cm,height=2.2cm]{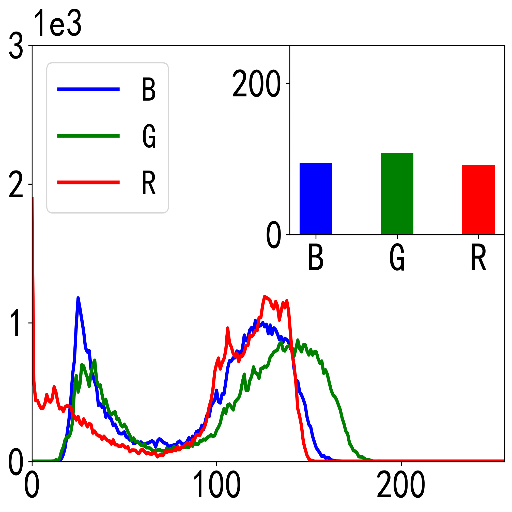}
    \includegraphics[width=2.8cm,height=2.2cm]{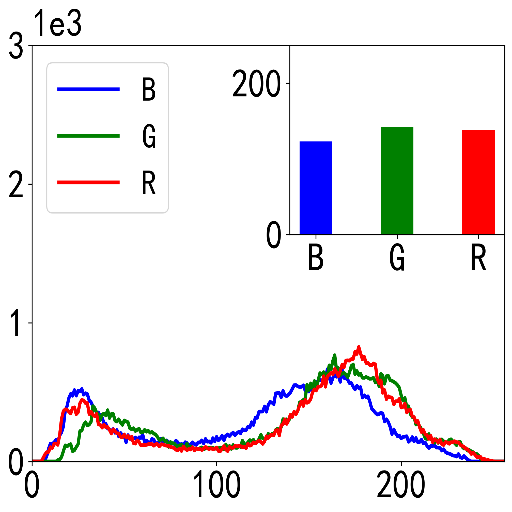}
    \includegraphics[width=2.8cm,height=2.2cm]{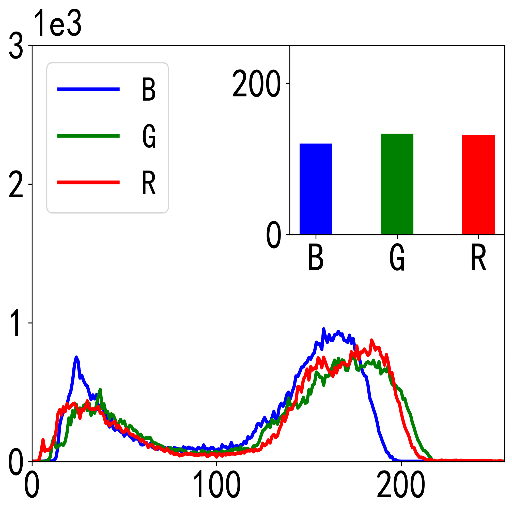}
    \includegraphics[width=2.8cm,height=2.2cm]{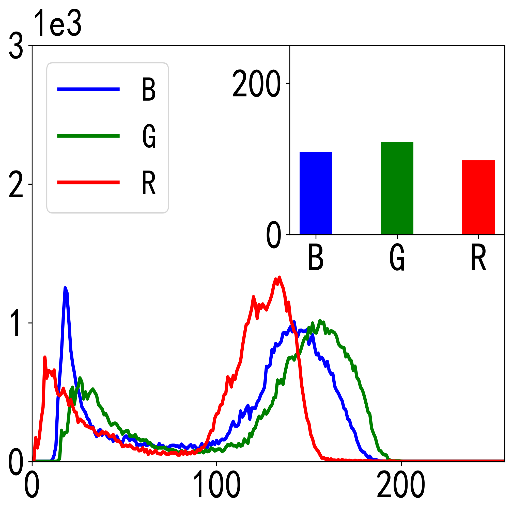}

    \leftline{\hspace{1cm} Distortion \hspace{1.3cm} UWNet \hspace{1.3cm} PhysicalNN \hspace{1.0cm} FUnIEGAN \hspace{1.0cm} WaterNet \hspace{1.2cm} ADMNNet }

    \vspace{0.1cm}

    \includegraphics[width=2.8cm,height=2.2cm]{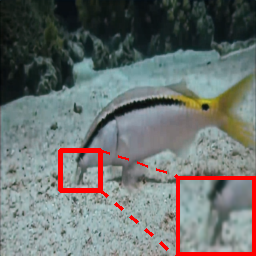}
    \includegraphics[width=2.8cm,height=2.2cm]{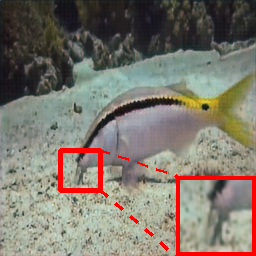}
    \includegraphics[width=2.8cm,height=2.2cm]{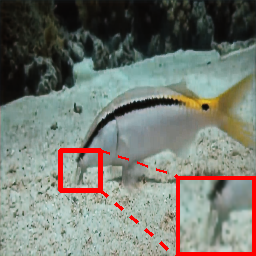}
    \includegraphics[width=2.8cm,height=2.2cm]{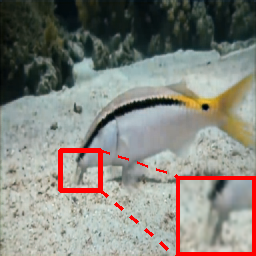}
    \includegraphics[width=2.8cm,height=2.2cm]{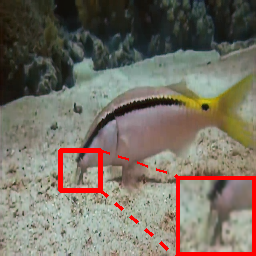}
    \includegraphics[width=2.8cm,height=2.2cm]{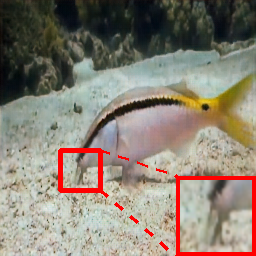}

    \includegraphics[width=2.8cm,height=2.2cm]{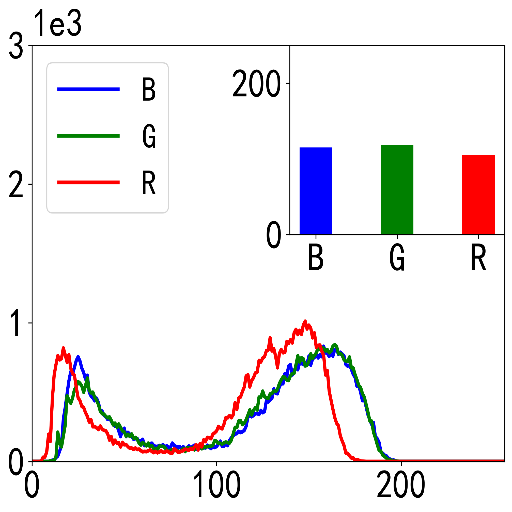}
    \includegraphics[width=2.8cm,height=2.2cm]{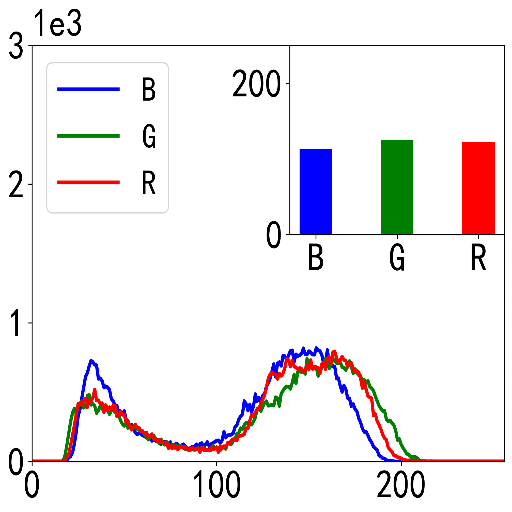}
    \includegraphics[width=2.8cm,height=2.2cm]{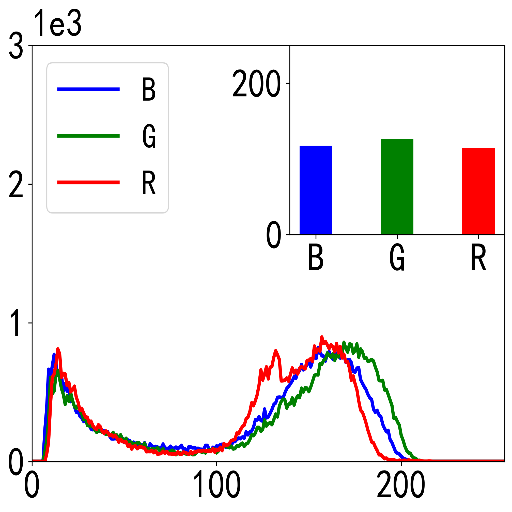}
    \includegraphics[width=2.8cm,height=2.2cm]{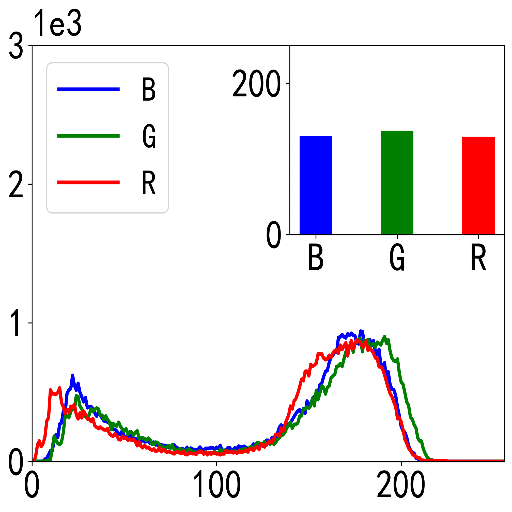}
    \includegraphics[width=2.8cm,height=2.2cm]{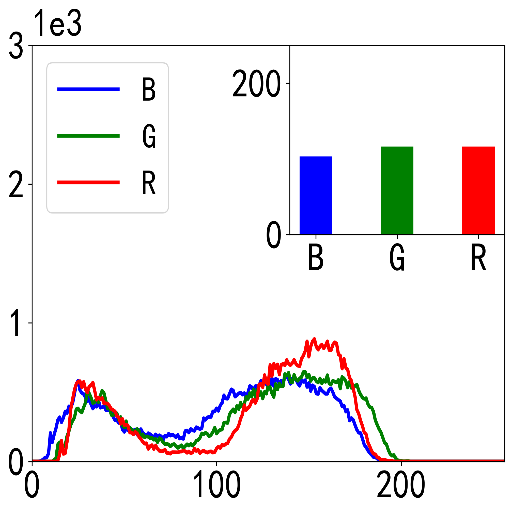}
    \includegraphics[width=2.8cm,height=2.2cm]{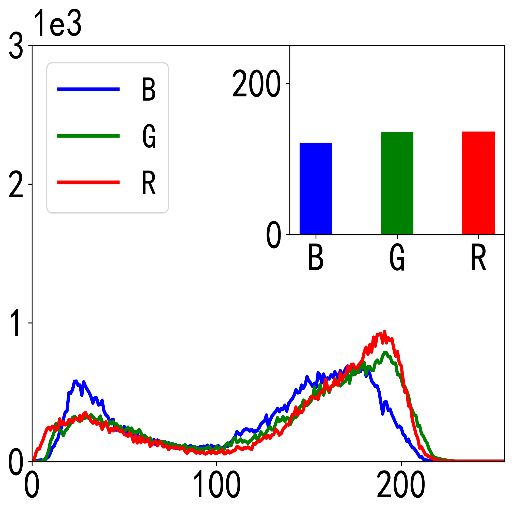}

    \leftline{\hspace{1.2cm} UColor \hspace{1.5cm} UGAN \hspace{1.5cm} SGUIE \hspace{1.7cm} UIE-WD \hspace{1.6cm} SCNet \hspace{1.7cm} STSC}
    
    \vspace{0.1cm}

    \includegraphics[width=2.8cm,height=2.2cm]{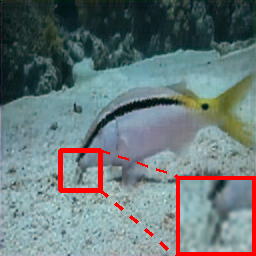}
    \includegraphics[width=2.8cm,height=2.2cm]{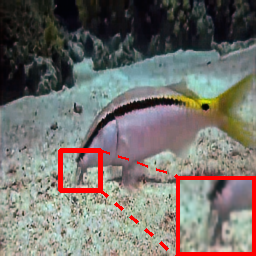}
    \includegraphics[width=2.8cm,height=2.2cm]{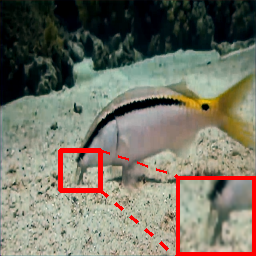}
    \includegraphics[width=2.8cm,height=2.2cm]{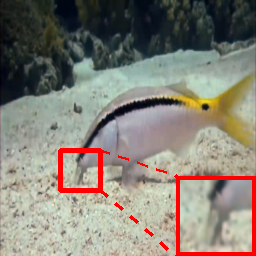}
    \includegraphics[width=2.8cm,height=2.2cm]{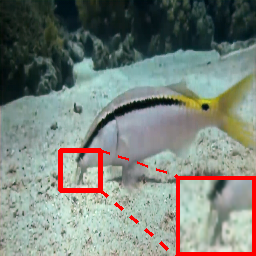}
    \includegraphics[width=2.8cm,height=2.2cm]{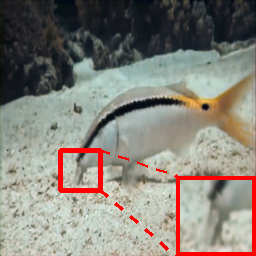}

    \includegraphics[width=2.8cm,height=2.2cm]{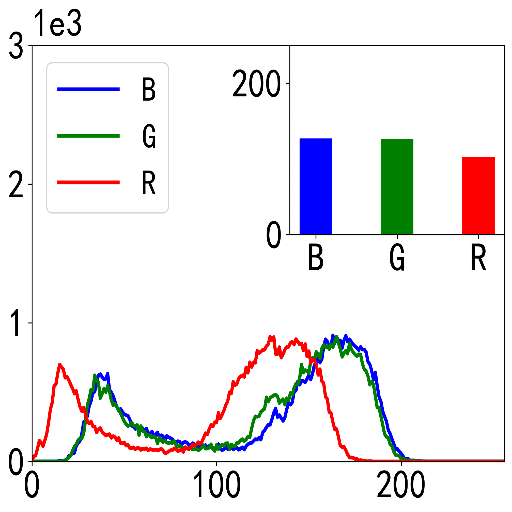}
    \includegraphics[width=2.8cm,height=2.2cm]{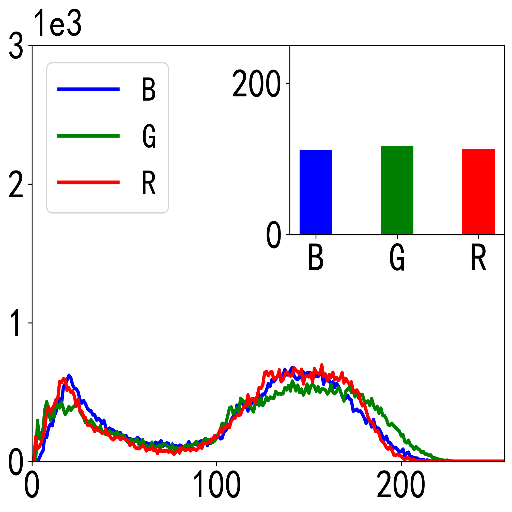}
    \includegraphics[width=2.8cm,height=2.2cm]{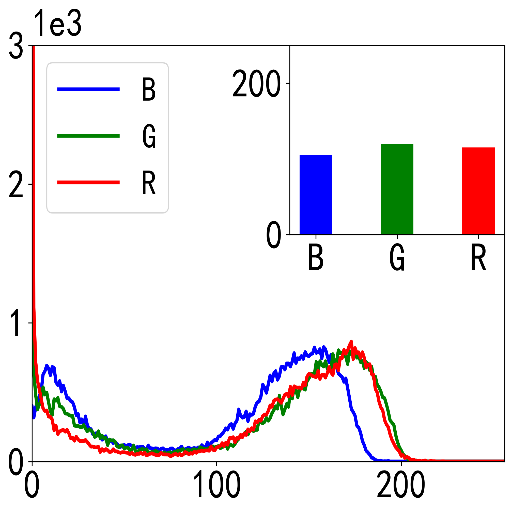}
    \includegraphics[width=2.8cm,height=2.2cm]{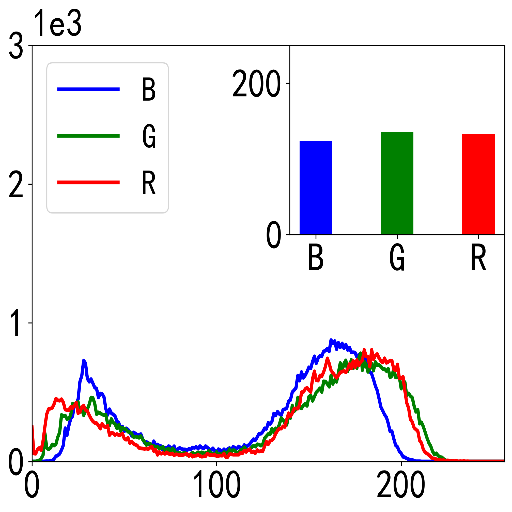}
    \includegraphics[width=2.8cm,height=2.2cm]{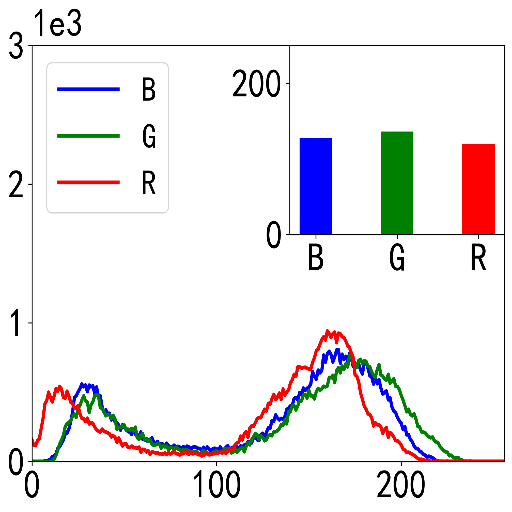}
    \includegraphics[width=2.8cm,height=2.2cm]{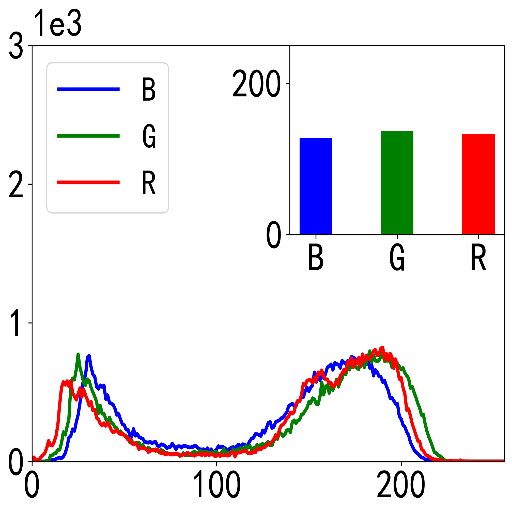}
    \leftline{\hspace{1.2cm} U-Trans \hspace{1.4cm} CECF \hspace{1.5cm} Semi-UIR \hspace{1.4cm} UIE-DM \hspace{1.3cm} HCLR-Net \hspace{1.2cm} GUPDM}

    \caption{Visual results obtained on challenge-60. The curve figure denotes the pixel histogram. The bar figure gives the average pixel value of each channel.}
    \label{fig:visual_results_challenge60}
  \end{figure*}

\subsection{Disentanglement $\&$ Fusion}
\label{subsec:disentanglement_and_fusion}

\subsubsection{Physical Embedding}
Neural network-based UIE methods can often achieve impressive performance on specific datasets, but the generalization capabilities and interpretability they exhibit may be limited. Therefore, physical models are integrated into the data-driven training process by various algorithms. IPMGAN \cite{liu2021ipmgan} embeds the Akkaynak-Treibitz model \cite{akkaynak2018revised} introduced in Section \ref{subsec:physical_models} into a generative adversarial network. The desired reference image can be refered as 
\begin{equation}
    y = \frac{x - B^{\infty}\left(1 - e^{-\beta^{B}(v_{B}) d}\right)}{e^{-\beta^{D}(v_{D}) d}}.
\end{equation}

This physical representation can be reformulated for obtaining enhanced image $\hat{y}$ as
\begin{equation}
    \hat{y} = \frac{x - \widehat{B}^{\infty}(1 - \widehat{S})}{\widehat{T}},
\end{equation}
where $\widehat{T}$ and $\widehat{S}$ are the estimated $e^{-\beta^{D}(v_{D})z}$ and $e^{-\beta^{B}(v_{B})d}$, respectively. The $\widehat{B}^{\infty}$ denotes the estimated veiling light. Through weight-sharing physical parameters ($\widehat{T}$, $\widehat{S}$ and $\widehat{B}^{\infty}$) estimation, the enhanced image can be derived from the inverse imaging model. The distortion process of the underwater image is considered as a horizontal and vertical process by ACPAB \cite{lin2020attenuation}. From a horizontal perspective, the parameters required by the physical model, namely attenuation coefficient, transmission map and background light, are estimated by the neural network. From a vertical perspective, an attenuation coefficient prior is embedded into the enhancement network. USUIR \cite{fu2022unsupervised} builds a physically decoupled closed-loop system. Three physical parameters are estimated by the network and priori assumptions. Then, the original attenuation image is synthesized inversely through the physical parameters. Through such a self-feedback mechanism, the training of USUIR can be carried out in an unsupervised manner. It is worth pointing out that the modeling of the imaging process of the underwater environment is still a task under exploration, and no perfect model has yet been proposed.

\subsubsection{Retinex Model}
According to the Retinex model \cite{qi2023deep}, the image can be decomposed into the Hadamard product of two components: reflection and illumination. The form of the Retinex model is simpler than the physical models commonly used in the UIE task since it requires fewer parameters to be estimated. The multi-scale Retinex is employed by CCMSR-Net \cite{qi2023deep}, which is
\begin{equation}
    R = \sum_{n=1}^{N} w_{n}[\log{x} - \log{x \ast G_{n}}],
\end{equation}
where $n$ and $w_{n}$ represent the scale index and corresponding weight factor, respectively. $G_{n}$ stand for the Gaussian kernel under scale $n$. The $R$ denote the reflectance. CCMSR-Net decomposes the enhancement task into two procedures. A subnetwork for color correction and a multiscale Retinex subnetwork for estimating the illumination, which take into account two consecutive procedures, are integrated by CCMSR-Net. ReX-Net \cite{zhang2023rex} uses two encoders, namely the original image encoder and the Retinex-based reflectance encoder, to extract complementary content and color information. Based on the self-information extracted which utilizes the Retinex decomposition consistency, ASSU-Net \cite{khan2023underwater} designs a pipeline to improve the contrast of the weak and backlit areas.

\subsubsection{Color Space Fusion} 
Different attributes are held by different color spaces. RGB representation is the most commonly used color space for the UIE task. The three color channels of RGB space are highly correlated. This property makes it possible for the RGB space to be easily affected by the changes in luminance, occlusion, and shadow \cite{wang2021uiec}. Therefore, color spaces with other properties are explored by UIE algorithms. A HSV global-adjustable module is designed in UIEC${}^2$-Net \cite{wang2021uiec}, which can be used for adjusting the luminance, color and saturation by utilizing a piece-wise linear scaling curve layer. Meanwhile, a HSV loss $\mathcal{L}_{hsv}$ is adopted by UIEC${}^2$-Net as
\begin{equation}
    \mathcal{L}_{hsv} = ||\widehat{S}\widehat{V}\cos{\widehat{H}} - {SV}\cos{H}||_{1},
\end{equation}
where $\widehat{H}$, $\widehat{S}$ and $\widehat{V}$ denote the predictions of $H \in [0, 2\pi)$, $S \in [0, 1]$ and $C \in [0, 1]$, respectively. MTNet \cite{moran2023mtnet} proposes a loss function measured from HSV space to better restore contrast and saturation information. An RGB-HSV dual-color space-guided color estimation block is proposed by UGIF-Net \cite{zhou2023ugif} to generate comprehensive color information. The LAB space is explored by TCTL-Net \cite{li2023tctl} and P2CNet \cite{rao2023deep} to improve color recovery performance. JLCL-Net \cite{xue2021joint} converts the distorted image and reference image into YCbCr space to improve the sensitivity of luminance and chrominance. RGB, HSV and LAB are simultaneously embedded into the encoding path by UColor \cite{li2021underwater} to construct features with a unified representation.

\subsubsection{Water Type Focus} 
Due to the complexity of underwater scene imaging, underwater distorted images may exhibit different properties. Common imaging environments include shallow coastal waters, deep oceanic waters, and muddy waters. The lighting conditions of different underwater environments are diverse. Therefore, a challenging topic is how to handle different water types using a single UIE model. Aiming at learning water type's desensitized features, SCNet \cite{fu2022underwater} performs normalization operations in both spatial and channel dimensions. An instance whitening is designed in the encoding-encoding structure of SCNet. For the $n$-th example $x_{n} \in \mathbb{R}^{C \times HW}$ in a mini-batch, the designed instance whitening $\Gamma(\cdot)$ can be expressed as
\begin{equation}
    \Gamma(x_{n}) = \Sigma^{-1/2}(x_{n} - \mu)\gamma + \beta,
\end{equation}
where $\gamma$ and $\beta$ denote scale and shift which can be dynamically learned. The mean vector $\mu$ and covariance matrix $\Sigma$ are computed with each individual example by
\begin{equation}
    \begin{cases}
        \mu = \frac{1}{HW} x_{n}, \\
        \Sigma = \frac{1}{HW} (x_{n} - \mu)(x_{n} - \mu)^{T} + \alpha{I}, \\
    \end{cases}
\end{equation}
where $\alpha$ and $I$ denote a small positive number and identity matrix, respectively. By using the instance whitening operation, the influence of diverse water types can be reduced. An adversarial process involving latent variable analysis is used by DAL \cite{uplavikar2019all} to disentangle the unwanted nuisances corresponding to water types. A nuisance classifier is designed by DAL, which classifies the water type of the distorted image according to its latent vector. IACC \cite{zhou2024iacc} designs an underwater convolution module that can learn channel-specific features and adapt to diverse underwater environments by leveraging the mini-batch insensitivity of instance normalization.

\subsubsection{Multi-input Fusion}
Research \cite{li2019underwater} shows that preprocessing input can be beneficial for the UIE task. WaterNet \cite{li2019underwater} applies White Balance (WB), Histogram Equalization (HE) and Gamma Correction (GC) algorithms to distorted underwater images. The WB may adjust the color distortion. The HE and GC can increase contrast and optimize dark regions. A gated fusion approach is used by WaterNet to obtain an integrated output. The gated enhanced result is
\begin{equation}
    \hat{y} = R_{WB} \odot C_{WB} + R_{HE} \odot C_{HE} + R_{GC} \odot C_{GC},
\end{equation}
where $R_{WB}$, $R_{HE}$ and $R_{GC}$ mean the refined images obtained by the corresponding pre-process methods. The $C_{WB}$, $C_{HE}$ and $C_{GC}$ denote the learned confidence maps. The $\odot$ is element-wise production. MFEF \cite{zhou2023multi} and F2UIE \cite{verma2023f2uie} utilize the WB and Contrast-limited Adaptive Histogram Equalization (CLAHE) to obtain high-quality input with better contrast.

\section{Experiments}
\label{sec:experiments}

To promote research on the UIE task, we provide a fair evaluation of UIE algorithms that have been proven to be effective on benchmark underwater datasets.

\subsection{Settings}

\noindent \textbf{Algorithms.} The DL-UIE algorithms include UWNet \cite{naik2021shallow}, PhysicalNN \cite{chen2021underwater}, FUnIEGAN \cite{islam2020fast}, WaterNet \cite{li2019underwater}, ADMNNet \cite{yan2022attention}, UColor \cite{li2021underwater}, UGAN \cite{fabbri2018enhancing}, SGUIE \cite{qi2022sguie}, UIE-WD \cite{ma2022wavelet}, SCNet \cite{fu2022underwater}, STSC \cite{wang2022semantic}, U-Trans \cite{peng2023u}, CECF \cite{cong2024underwater}, Semi-UIR \cite{huang2023contrastive}, UIE-DM \cite{tang2023underwater}, HCLR-Net \cite{zhou2024hclr} and GUPDM \cite{mu2023generalized}. They all have open source code implemented in PyTorch.

\noindent \textbf{Metrics.} The full-reference metrics include PSNR and SSIM. The no-reference metrics include UIQM \cite{panetta2015human}, UCIQE \cite{yang2015underwater} and URANKER \cite{guo2023underwater}. For all experimental analyses and discussions below, we assume that the quantitative evaluation metrics that are widely used in existing literature are reliable. Otherwise, we can not draw any conclusions.

\noindent \textbf{Datasets.} The benchmark datasets with paired images include UIEB \cite{li2019underwater}, LSUI \cite{peng2023u}, \{EUVP-D/EUVP-I/EUVP-S\} from EUVP \cite{islam2020fast} and UFO-120 \cite{islam2020simultaneous}. The benchmark datasets without references include Challenge-60 from UIEB \cite{li2019underwater}, U45 \cite{li2019fusion}, UCCS from RUIE \cite{liu2020real}, and EUVP-330 from EUVP \cite{islam2020fast}. Different semi-supervised algorithms require different types of external data. In order to ensure fairness as much as possible, we use the pretrained models on the UIEB (full-reference training set) dataset for the performance test of no-reference datasets.


\noindent \textbf{Factor Settings.} In order to do our best to keep the comparative experiments fair, we have unified the following factors that may affect the experimental results.
\begin{itemize}
    \item The batch size is set to 8.
    \item The image size in both training and testing phases is set to $256 \times 256$.
    \item Three data augmentation methods are used, namely horizontal random flipping, vertical random flipping and random cropping.
    \item The feature extraction part of any model does not use pre-trained patterns guided by external data.
    \item The partition ratio of training and testing data may be different in existing literature. Here we use a uniform ratio to divide the training and test data.
    \item Different models may require different initial learning rates, learning rate decay strategies, and optimizers. We follow the settings given in their respective papers. We did not conduct additional hyperparameter searches for each model.
    \item In order to prevent differences in metric values caused by different toolboxes or implementations \cite{chen2023towards}, we unified the code for metric calculations. 
\end{itemize}

\subsection{Comparison among DL-UIE Algorithms}

\noindent \textbf{Evaluation of Fitting Ability on Full-Reference Benchmark Datasets.} Table \ref{tab:full_reference_results} shows the results obtained by various DL-UIE algorithms on datasets with paired data. Overall, the quantitative evaluation metrics obtained by UIE-DM are the best. However, on most datasets, there is no significant difference in the quantitative evaluation results (i.e., PSNR and SSIM) obtained by the top 5 algorithms. What is particularly noteworthy is that for EUVP-D, EUVP-I and EUVP-S, the PSNR and SSIM of the top 10 algorithms are approaching the same level. The visualization results in Fig. \ref{fig:visual_results_uieb_2774} also illustrate that multiple DL-UIE algorithms achieve similar visual effects. Not only that, the color histogram also shows that the results obtained by different algorithms are close in terms of pixel statistics. For these UIE algorithms, the representation capabilities of the neural networks designed with different architectures may be very close. This phenomenon implies that the pursuit of higher PSNR and SSIM may be difficult. 

\noindent \textbf{Evaluation of Generalization Ability on No-Reference Benchmark Datasets.} We tested two kinds no-reference metrics. The first is the manually calculated metrics UIQM and UCIQE, which are shown in Table \ref{tab:non_reference_results}. The UIQM and UCIQE values of different DL-UIE algorithms given in Table \ref{tab:non_reference_results} show that UGAN achieves the overall best performance. Meanwhile, in terms of the evaluation of generalization ability, the UIQM and UCIQE values obtained by different algorithms are close. The second is URANKER obtained by data-driven training, which is given in Fig. \ref{fig:nriqa}. The best URANKER values are obtained by M (UIE-DM), K (CECF), E (UGAN), and I (STSC), respectively. The second best URANKER values are obtained by K (CECF), I (STSC), H (SCNet), M (UIE-DM) respectively. For the evaluation of URANKER, UIE-DM has the best overall evaluation effect. Considering Table \ref{tab:non_reference_results} and Fig. \ref{fig:nriqa} simultaneously, the generalization ability of UGAN is the best. Fig. \ref{fig:visual_results_challenge60} shows the visual enhancement results of the evaluation of generalization ability. There are obvious differences in the visual results and pixel histograms of images obtained by different algorithms. Some enhanced images still show blue-green of green effect.

\noindent \textbf{Overall Conclusions.} Under our experimental settings, UIE-DM and UGAN achieved the best performance in fitting ability on full-reference datasets and generalization ability on no-reference datasets, respectively.

\section{Future Work}
\label{sec:future_work}
According to our discussion of existing progress and experiments, there are still challenging problems that remain unsolved. We raise the following issues worthy of study.

\begin{itemize}
    \item \textbf{Towards high-quality pairwise data synthesis by game engines.} Constructing a large-scale database contains real-world paired underwater images is almost impossible. Existing methods synthesizing data by algorithms are difficult to accurately simulate different influencing factors, such as illumination intensity, number of suspended particles, water depth and scene content. For game engines that can customize extended functions, controlling influencing factors is an inherent advantage. For example, Liu et al. \cite{liu2023nighthazeformer} use the UNREAL game engine to simulate non-uniform lighting, low-illumination, multiple light sources, diverse scene content and hazy effects at nighttime. It is worth exploring how game engines can be used to build datasets for the UIE task.
    \item \textbf{Effects on downstream vision tasks.} High-level vision tasks, or be called downstream tasks of the UIE task, have been used by existing research as a strategy to evaluate the performance of the UIE model itself. An intuitive and widely adopted hypothesis is that the UIE model's ability to enhance the distorted images is positively related to its ability to facilitate downstream tasks. However, a recent study \cite{wang2023underwater} on the object detection task and the UIE task discovered a surprising phenomenon, and here we directly quote their conclusion, ``One of the most significant findings is that underwater image enhancement suppresses the performance of object detection.''. A comprehensive research on the correlation between the performance of the UIE and downstream tasks may provide a reliable conclusion.
    \item \textbf{Cooperation with large-scale pretrained vision-language models.} The texture and local semantics contained in underwater images have been extensively studied. However, the global semantics that may be provided by language models have not been embedded into UIE models by existing research. Large-scale pretrained vision-language models, such as CLIP \cite{radford2021learning,zhang2024atlantis}, can provide human-level high-level semantic information. Text-image multi-modal restoration models have been partially studied in rain and haze removal \cite{luo2024controlling}. Language features may facilitate the performance of UIE models.
    \item \textbf{Non-uniform illumination.} Artificial light sources are used by underwater equipment when the depth of the water exceeds the range that can obtain a properly illuminated image. However, unlike the smoothness of natural light, the illumination of images captured under artificial light sources may be non-uniform \cite{hou2023non}. The enhancement task under non-uniform lighting is worth digging into.
    \item \textbf{Reliable evaluation metrics.} The difference between the enhanced result and the reference can be reliably evaluated by using full-reference evaluation metrics. However, obtaining ground-truth in the underwater imaging process is extremely challenging. Therefore, numerous literature use no-reference evaluation metrics to evaluate the effectiveness of their proposed algorithms. The reliability of no-reference metrics needs to be consistent with human subjective aesthetics. Currently, in-depth research on related topics is still scarce. More research about subjective and objective evaluations are worth conducting.
    \item \textbf{Combination with other image restoration tasks.} As we discuss in this paper, there are inherent differences between the UIE task and other image restoration tasks. Preliminary attempts \cite{li2022all} have been made to combine tasks such as haze, rain and noise removal. The low-level features learned by different in-air image restoration models may be beneficial to each other. Utilizing auxiliary knowledge provided by other restoration tasks may improve the performance of UIE models.
\end{itemize}


\section{Conclusion}
\label{sec:conclusion}
In this paper, we attempt to conduct a systematic review of the research of the underwater image enhancement task. We first investigated the research background and related work, which include physical models describing the degradation process, data construction strategies for model training, evaluation metrics implemented from different perspectives, and commonly used loss functions. We then provide a comprehensive taxonomy of existing UIE algorithms. According to the main contributions of each algorithm, state-of-the-art algorithms are discussed and analyzed from different perspectives. Further, we perform quantitative and qualitative evaluations on multiple benchmark datasets containing synthetic and real-world distorted underwater images. Finally, based on our summary of algorithms and analysis of the experiments, open issues and challenging topics are raised.

\normalem
\bibliography{ref.bib}


\end{document}